\documentclass[lettersize,journal]{IEEEtran}
\usepackage{amsmath,amsfonts}
\usepackage{algorithmic}
\usepackage{algorithm}
\usepackage{array}
\usepackage{booktabs}
\usepackage{multirow}
\usepackage{xcolor}
\usepackage{caption}
\usepackage{textcomp}
\usepackage{stfloats}
\usepackage{url}
\usepackage{verbatim}
\usepackage{graphicx}
\usepackage{cite}
\begin{document}

\title{DPA: Decoupling Product-Agnostic Anomaly Representations for\\Zero-shot Anomaly Generation}

\author{Hang Yao, Yansheng Fu, Ming Liu, Zifei Yan, Yanli Ji, Hongzhi Zhang, Wangmeng Zuo 
\thanks{Hang Yao, Yansheng Fu, Ming Liu, Zifei Yan, Hongzhi Zhang, Wangmeng Zuo are with the Faculty of Computing, Harbin Institute of Technology, Harbin, 150000, China, (E-mail:hyao\_1@outlook.com). (Corresponding author: Ming Liu)}
\thanks{Yanli Ji is with the School of Intelligent Systems Engineering, Sun Yat-Sen University, Guangzhou, 510000, China}}

\markboth{Journal of \LaTeX\ Class Files,~Vol.~14, No.~8, August~2021}%
{Shell \MakeLowercase{\textit{et al.}}: A Sample Article Using IEEEtran.cls for IEEE Journals}


\twocolumn[{%
	\renewcommand\twocolumn[1][]{#1}%
	\maketitle
	\begin{center}
		\centering
		\includegraphics[width=\linewidth]{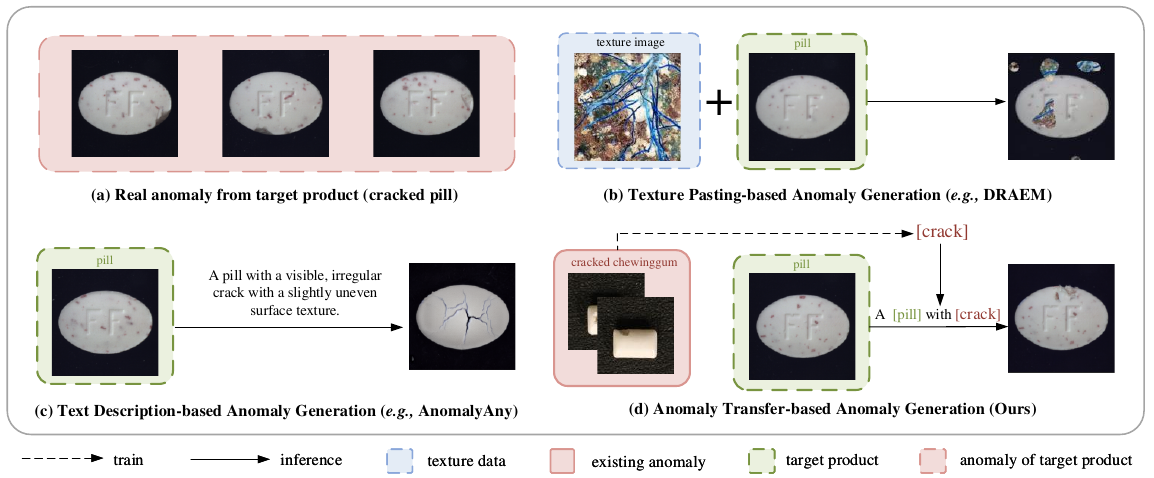}
		\vspace{-.5em}
		\captionof{figure}{Comparison of zero-shot anomaly generation setting with different anomaly source. (a) shows the real target-product anomalies. Texture pasting-based setting in (b) uses generic texture images as the anomaly source and often generate unrealistic anomaly patterns. Text description-based setting in (c) uses text descriptions as anomaly source. However, text cannot fully specify diverse anomaly appearances, and pre-trained anomaly concepts may diverge from real manufacturing anomalies. Our anomaly transfer-based setting in (d) instead learns a anomaly concept from real anomalies of other product and transfers it to a normal target image. This enables generating anomalies that closely resemble the real anomalies in (a).}
		\label{fig:setting}
	\end{center}
}]

\begin{abstract}
Industrial anomaly detection benefits from anomaly samples, yet newly deployed products typically provide only normal images, making anomaly samples difficult to collect. 
Zero-shot anomaly generation offers a promising solution which avoids collection of target-product anomalies. However, existing methods mainly rely on texture images or text descriptions as anomaly sources, which often produce unrealistic anomalies. Observing that similar  anomalies can recur across different products, we propose \emph{anomaly transfer-based zero-shot generation}, which reuses real anomalies from existing source products, making target-product anomalies no longer necessary to generate realistic anomalious samples for unseen target products. Since not every anomaly type suits the target product, an anomaly type filtering mechanism first selects plausible source types. To transfer selected anomaly, we propose DPA, a diffusion-based framework that decouples product-agnostic anomaly representations. Instead of directly extracting anomaly representations, DPA learns product-irrelevant anomaly embeddings through training with the mismatched data pair, enabling transferable anomaly concept learning across products. Furthermore, we design an adaptive mask-guided pipeline that leverages adaptive masks to control the positional and geometric plausibility of generated anomalies during generation. A training-free anomaly labeling module is further introduced to produce pixel-level annotations aligned with generated anomalies. Extensive experiments on MVTec-AD, VisA, and a dedicated anomaly-transfer benchmark demonstrate that the proposed setting and DPA generate more realistic anomalies and significantly improve downstream anomaly detection performance under both zero-shot and few-shot settings. Source code and models will be released.
\end{abstract}


\section{Introduction}
\label{sec:introduction}

\IEEEPARstart{I}{ndustrial} anomaly detection (IAD)~\cite{zhu2026anomaly,gao2026dual,zhang2026tunclip}, as a critical step of quality control, is drawing increasing attention in the intelligent industry era. Meanwhile, the rapid development of modern manufacturing has led to diversified demands and constant product iteration. When deploying IAD models ~\cite{gao2022cas,zou2018deepcrack,li2026accurate} to these continuously introduced new products, both normal and anomalous samples are required for training is typically. In practice, however, while normal samples are readily obtainable, anomalous samples are extremely scarce. To alleviate this data scarcity, a straightforward and practical solution is to generate sufficient and diverse anomalous samples using only normal data, \textit{i.e.}, zero-shot anomaly generation.

Despite their diversity, zero-shot anomaly generation methods follow the same general formulation: an \emph{anomaly source} is combined with a normal target image to produce a target anomalous image and its annotation. 
Within this formulation, the most critical element is the anomaly source, that is, the information that specifies what the anomaly should look like. The real target anomalies shown in Fig.\ref{fig:setting}(a) represent the desired output and, ideally, the most reliable anomaly source. However, they are inaccessible under zero-shot setting.
Early \emph{texture pasting-based methods}~\cite{zavrtanik2021draem,zhang2025diffusionad,yao2024glad} use external texture images as anomaly source and insert them into randomly sampled regions of normal images. Though efficient, they produce anomalies that are often unnatural and deviate significantly from real anomaly patterns, as illustrated in Fig.\ref{fig:setting}(b). Recent, \emph{text description-base methods}~\cite{sun2025unseen} use text descriptions as the source and prompt a pre-trained diffusion model (Fig.\ref{fig:setting}(c)). However, text cannot fully specify the appearance of diverse industrial anomalies, and the anomaly concepts learned by the pre-trained generation model diverge from real manufacturing anomalies.Therefore, this anomaly source still lack authenticity, resulting in a substantial gap in visual appearance between generated anomalies and the expected target anomalies.

Although anomalous data are difficult to collect for a newly deployed product, factories often retain real anomaly samples from existing production. Many anomaly concepts, such as scratches, cracks, discoloration, and bent components, recur across related product families and exhibit similar visual patterns. This cross-product commonality makes existing anomalies a more faithful anomaly source than generic textures or text descriptions, while avoiding the costly collection of target-product anomalies. Motivated by this observation, we propose \emph{anomaly transfer-based zero-shot generation}, illustrated in Fig.\ref{fig:setting}(d). In this setting, anomaly concepts extracted from real anomalies of source products are transferred to unseen target products, with no real target anomaly samples required.

However, not every source anomaly is meaningful for the given target product. For example, a bent-lead anomaly is applicable only to products containing leads, whereas transferring it to a bottle would create semantically invalid training data. We therefore introduce a product-aware anomaly type filtering mechanism before transfer. For each target product, semantically incompatible source anomaly types are discarded, and only reasonable types are retained for generation. In practice, we use the vision-language model QWen3-VL to make this compatibility decision before generation, after which the compatibility decision is cached and reused for all subsequent samples. Importantly, the vision-language model only decides whether an anomaly is suitable for the product. The anomaly concept itself is learned from the selected source anomaly.

\begin{figure}
	\centering
	\includegraphics[width=\linewidth]{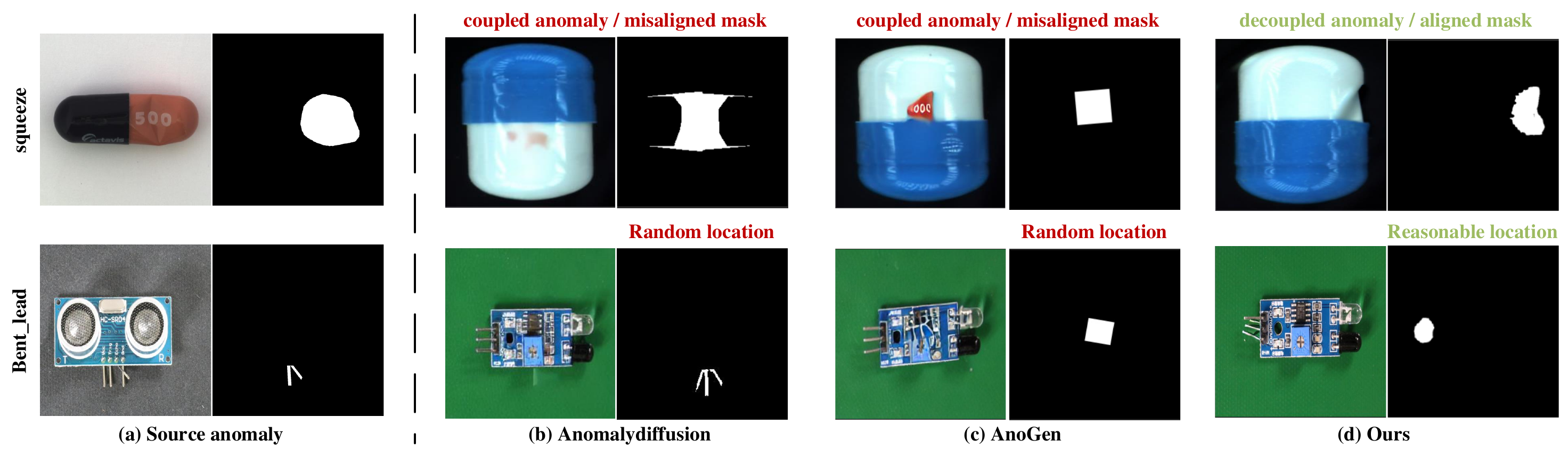}
	\caption{Results of direct cross-product anomaly transfer and the proposed DPA. (a) A source anomalous image contains both anomaly and source-product appearance. Directly applying existing anomaly generators to cross-product transfer, as illustrated by (b) AnomalyDiffusion~\cite{hu2024anomalydiffusion} and (c) AnoGen~\cite{gui2024few}, may transfer source-product information together with the anomaly and produce randomly located or misaligned masks. (d) DPA transfers a decoupled anomaly concept to a reasonable location of the target product and produces a mask aligned with the generated anomaly.}
	\label{fig:intro_generation}
\end{figure}

A straightforward approach to achieve the transfer of selected anomaly types, is to use existing anomaly generation methods~\cite{hu2024anomalydiffusion,gui2024few} to extract anomaly representations from the source anomaly and apply it to a normal target-product image \footnote{Although these methods are few-shot methods that require target anomaly data, replacing target anomalies with source anomalies makes them formally applicable to anomaly transfer.}. However, such direct adaptation either fails to inject the anomaly into the target image or produces obvious artifacts, as shown in Fig.\ref{fig:intro_generation}(b) and (c). We attribute such failures to the impure anomaly representations extracted with these methods, where the vanilla feature extraction module inevitably couples anomaly representations with product representations.

This entanglement motivates our \emph{Anomaly Concept Decoupling} (ACD). First, ACD introduces two learnable embeddings: a product embedding shared across normal and anomalous samples to represent product appearance, and an anomaly embedding dedicated to anomaly semantics. And then, to assign complementary responsibilities to these embeddings, we construct three types of input-target pairs for embeddings training: masked anomaly $\rightarrow$ masked anomaly, anomaly $\rightarrow$ normal, and normal $\rightarrow$ anomaly. Guided by the corresponding embeddings, these pairs require the frozen diffusion model to perform three tasks: preserving, removing, and inserting anomalies, respectively. The preserve pair directly learns from the anomalous region to capture visual anomaly appearance, whereas the remove and insert pairs encourage the embedding to understand abstract anomaly concepts by erasing and adding anomalies. Together, these complementary tasks encourage it to decouple an abstract and transferable anomaly concept, rather than memorizing source-product appearance.

Another problem of these generation methods is the unreasonable mask, which is also shown in Fig.\ref{fig:intro_generation}(b) and (c). Existing methods typically use predefined masks as both generation constraints and ground-truth annotations. On the one hand, the masks are generated via random algorithms, which means that the mask regions may not be the place where a specific anomaly occurs. On the other hand, the generated anomalies are often unaligned with the masks (typically a circle or square), leading to imprecise annotations.

To address the mask issues, we propose an adaptive mask-guided pipeline comprising three modules: \emph{Initial Mask Generation} (IMG), \emph{Mask Refinement}(MR), and \emph{Anomaly Labeling}(AL), which operate before, during, and after generation, respectively. Before generation, IMG determines the anomaly location via attention maps between anomaly embeddings and normal image features, and sets the size by sampling from the prior size distribution of existing anomalies, yielding a circular coarse mask as an initial estimate. During generation, MR dynamically refines this mask, aligning it with the evolving anomaly and guiding the anomaly shape toward more plausible forms. After generation, to further produce a high-resolution mask, AL incorporates the refined masks, attention maps, and multi-scale VAE decoder features of normal and anomalous samples, to produce pixel-level precise annotations. These three modules jointly ensure the generated anomalies have reasonable spatial locations and shapes with accurate labels.

To verify the effectiveness of our anomaly transfer-based generation setting and generation pipeline, extensive experiments are conducted on two widely-used datasets (\textit{i.e.}, MVTec-AD~\cite{MVTec-AD28} and VisA~\cite{VisA}, where anomaly embeddings are extracted from one of these two datasets and evaluated on the other one. We also integrate a dedicated dataset for anomaly transfer, comprising both source data for extracting anomaly and target data for anomaly generation. 
Furthermore, we evaluate our model under the few-shot anomaly generation setting (where a small number of target anomaly samples are allowed), demonstrating the generality and effectiveness of our anomaly generation method. The contributions of this paper are summarized as follows.

\begin{itemize}
	\item We unify zero-shot anomaly generation as the composition of an anomaly source and a normal target image, and formulate anomaly transfer-based zero-shot generation setting, which uses real anomalies from existing products as a more faithful source. 
	\item A anomaly concept decoupling method is proposed, which decouples product appearance and anomaly concept to two learnable embeddings through complementary preserve, remove, and add tasks. 
	\item We introduce an adaptive mask-guided generation and labeling pipeline which considers location and shape of generated anomalies, and produces high‑precision annotations aligned with the generated anomalies.
	\item Experiments on three datasets, multiple downstream detection methods, and both zero-shot and few-shot settings consistently validate the proposed setting and method.
\end{itemize}

\section{Related Work}
\label{sec:relatedwork}
\subsection{Industrial Anomaly Detection}
Due to the scarcity of anomaly data, mainstream IAD algorithms adopt an unsupervised paradigm~\cite{yao2024glad,yang2025padnet,gao2026dual,yang2026open}, relying solely on normal samples for detection. Some of these methods design complex network architectures to fit the normal data distribution and treat anomalies as outliers~\cite{DDAD8,guo2025dinomaly,luo2025exploring,Padim13,PatchCore24}. Although these algorithms achieve good performance, their complex network structures lead to slow inference speed. In contrast to these architecture-oriented approaches, another line of unsupervised IAD focuses on data synthesis~\cite{zavrtanik2021draem,li2024target,chen2024unified}. These methods design simple anomaly synthesis strategies and combine them with lightweight detection networks, achieving decent detection performance. Specifically, they mainly paste texture data onto normal samples to simulate anomalies. These methods are highly efficient and well-suited for practical production deployment. However, due to the significant gap between synthetic anomalies and real anomalies, there is room for further performance improvement. Our goal is to design a zero-shot anomaly generation method that produces more realistic anomalies than the aforementioned synthesis strategies. By using the generated anomalies to train such lightweight detection models, we aim to maintain their low computational cost while further improving detection performance, enabling fast and accurate anomaly detection.

\subsection{Diffusion model}
Based on principles of nonequilibrium thermodynamics ~\cite{15}, diffusion model (DM) ~\cite{DDPM16,nichol2021improved,dhariwal2021diffusion} is proposed for image generation, and utilized in a variety of downstream tasks ~\cite{smartcontrol,elite,controlvideo,videoelevator}. Denoising diffusion implicit model (DDIM)~\cite{DDIM17} considers the reverse process of DM as a non-Markovian process, which speeds up the inference greatly. The latent diffusion model (LDM)~\cite{SD30} conducts training and inference in the latent space, further reducing the resource and time costs. Recently, Text inversion~\cite{TextInversion35} and Dreambooth~\cite{Dreambooth36} learn the concept of subjects in a given reference set and synthesize novel renditions of them in different contexts for customized generation. However, these methods heavily rely on real anomaly data, requiring the collection and annotation of target-specific anomalous samples, which significantly hinders the practical deployment of anomaly generation models.

\subsection{Anomaly Generation}
The scarcity of anomaly data has drawn considerable research attention, prompting the development of various anomaly generation methods to address this challenge. Conventional methods ~\cite{zavrtanik2021draem,zhang2024realnet,chen2024unified} do not rely on target anomaly data and can be regarded as zero-shot anomaly generation approaches. These methods simulate anomalies by replicating normal regions, texture patches, or adding noise perturbations. Although capable of synthesizing anomalies, they suffer from limited realism. Inspired by diffusion-based customized generation approaches ~\cite{TextInversion35, Dreambooth36}, few-shot anomaly generation methods \cite{hu2024anomalydiffusion,gui2024few,dai2025seas,jin2025dual,xu2025training} learn anomaly concepts from target anomaly samples to generate anomalies with enhanced realism. However, such methods require prior collection and annotation of anomalous samples, which is difficult under the premise of anomaly data scarcity. Recent zero-shot anomaly generation methods ~\cite{sun2025unseen,lai2025anomalypainter} employ detailed textual descriptions to guide pretrained generation models for zero-shot anomaly generation. Nevertheless, for common pretrained generation models, the textual descriptions of certain anomaly concepts often misalign with the desired visual characteristics. In summary, existing methods struggle to simultaneously achieve realistic anomaly generation and freedom from collecting target anomaly data, leading to significant limitations in practical applications.

\section{Methodology}
\label{sec:method}

We first define a common paradigm for zero-shot anomaly generation:
\begin{equation}
	\underbrace{\mathcal{S}}_{\text{anomaly source}}+
	\underbrace{I_p^n}_{\text{target normal}}
	\xrightarrow{\mathcal{G}}
	(\hat I_p^a,\hat m_p^a),
	\label{eq:zero-shot-paradigm}
\end{equation}
where $\mathcal{S}$ provides the anomaly prior, $I_p^n$ is a normal image of target product $p$, and $\mathcal{G}$ produces a target anomalous image $\hat I_p^a$ and its label $\hat m_p^a$. Texture-pasting methods instantiate the paradigm as $\mathcal{G}(\mathcal{T},I_p^n)$, where $\mathcal{T}$ is a generic texture image; text-guided methods use $\mathcal{G}(d^a,I_p^n)$, where $d^a$ is an anomaly description. DPA instead uses $\mathcal{G}(\mathcal{D}_s^a,I_p^n)$, where $\mathcal{D}_s^a$ contains real anomalous images, providing a substantially more faithful anomaly source.

Accordingly, DPA follows a training-to-inference pipeline. During training, \textit{Anomaly Concept Decoupling} (ACD) learns a decoupled anomaly embedding from real source anomalies, by trained with the preserve, remove, and add training pairs. Once the anomaly representations have been decoupled, inference begins with product-aware anomaly type filtering. For each target product, DPA evaluates the available source anomaly types and retains only those that are semantically compatible with the product. For every retained anomaly-product pair, the decoupled anomaly is combined with the target-product to construct a joint text prompt, \textit{e.g.}, ``A [bent\_lead] [pcb3].'' This prompt is encoded as learned embeddings, and conditions the frozen diffusion model to transfer the selected anomaly to a target normal image. The subsequent \textit{Adaptive Mask-Guided Generation} determines a plausible location and size and progressively controls the anomaly shape with adaptive masks. After generation, a training-free \textit{Anomaly Labeling} converts the guided mask into an aligned high-resolution mask.

\subsection{Anomaly Concept Decoupling}

\begin{figure}[t]
	\centering
	\includegraphics[pagebox=cropbox,width=.99\linewidth]{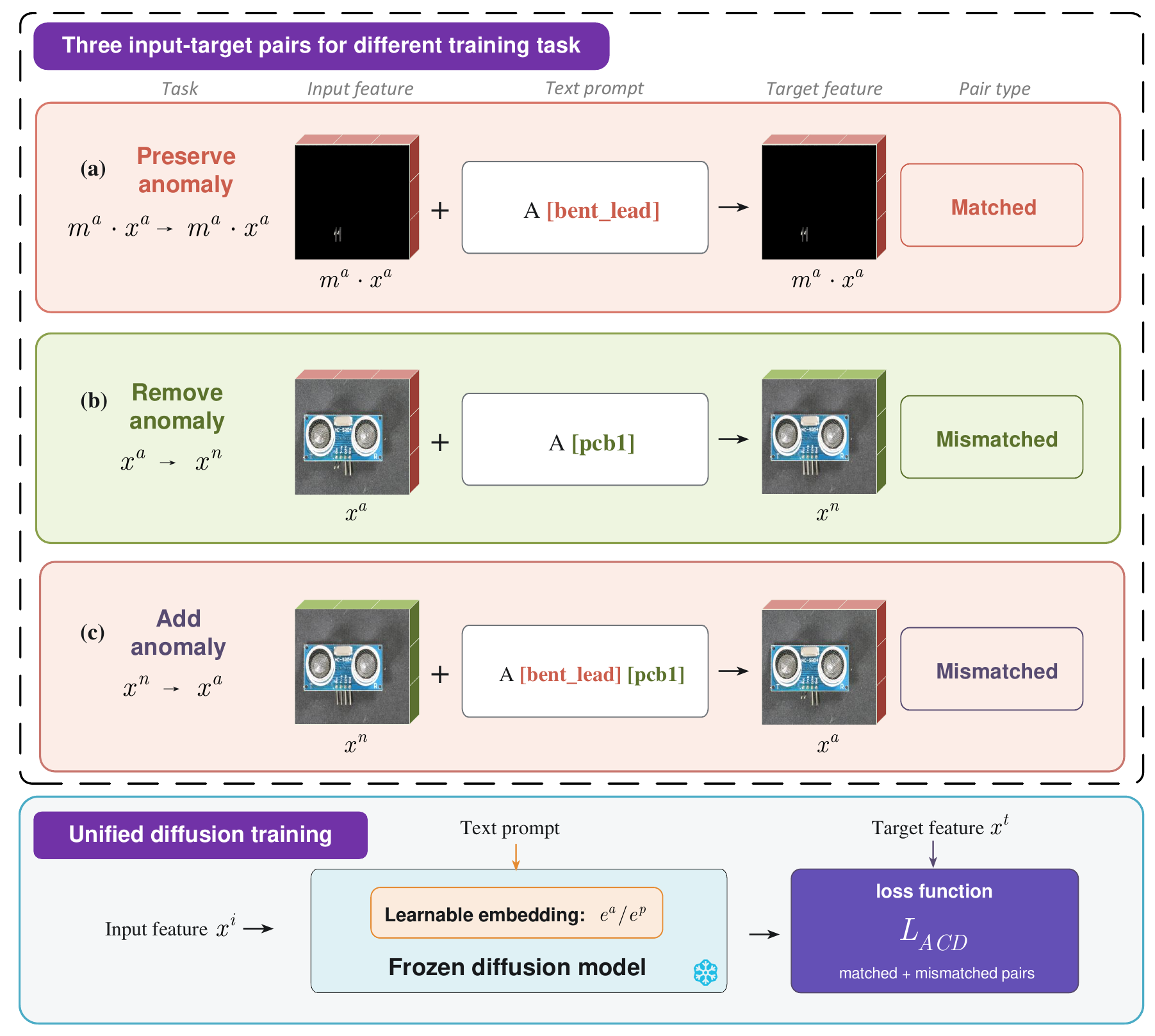}
	\vspace{-.5em}
	\caption{Decoupling training with three input--target pairs. The preserve, remove, and add tasks force the anomaly embeddings to learn pure anomaly concept form product appearance.}
	\vspace{-.5em}
	\label{fig:input}
\end{figure}

Although a real source anomaly provides a faithful visual prior in \ref{eq:zero-shot-paradigm}, it contains two inseparable factors: the source product and its anomaly. To obtain the product‑irrelevant anomaly concept, we decouple the anomaly concept from product information. The design of the training strategy is illustrated in Fig.\ref{fig:input}. We first introduce two learnable embeddings: a product embedding shared by both normal and anomalous samples, alongside a dedicated anomaly embedding activated only for abnormal samples. Guided by different embeddings, the diffusion model is required to complete three tasks by training with intentionally mismatched input-target pairs: retaining the anomaly, removing real anomalies and adding non-existent ones. This training strategy enhances the semantic representation of the anomaly embedding beyond pixel-level reconstruction, effectively decoupling anomaly patterns from product-specific information.

\subsubsection{\bf Decoupling Training Strategy.}
Given a spatially aligned anomalous--normal pair $(I^a,I^n)$ (the normal counterpart $I^n$ is obtained by completing the masked anomalous region with inpainting model~\cite{SD30}), we encode them as latent features $x^a$ and $x^n$ using a pre-trained VAE. Let $e^p$ and $e^a$ denote the learnable product and anomaly embeddings, respectively. As shown in Fig.\ref{fig:input}, the input feature $x^{i}$, target feature $x^{t}$, and conditioning embedding $e$ are sampled from\footnote{For training stability, conventional matched pairs, \textit{i.e.}, $(x^a,x^a)$ or $(x^n,x^n)$, are used with a probability of 50\%; the equation lists the proposed configurations.}
\begin{equation}
	\begin{split}
		(x^{i}, x^t, e) \in \{(&m^a\cdot x^a, m^a\cdot x^a, e^a),\\(&x^a, x^n, e^p), (x^n, x^a, [e^p, e^a])\},
	\end{split}
\end{equation}
where $m^a$ is the source anomaly mask. The three configurations respectively require the model to (1) preserve the masked anomaly pattern under $e^a$, (2) remove the anomaly from $x^a$ under the product embedding $e^p$, and (3) add the anomaly to $x^n$ under the joint condition $[e^p,e^a]$. Because the add and remove tasks share the same product embedding, this construction encourages source-product information to be captured by $e^p$ and the anomaly concept by $e^a$. 

\subsubsection{\bf Diffusion loss for mismatched pairs}
The standard diffusion objective assumes identical input and target features and therefore does not directly provide the correction required by the add and remove pairs above. We adopt the Anomaly-oriented Training Paradigm (ATP) loss~\cite{yao2024glad} for training, which supports reconstruction toward a different target. Let $x^i_t$ be the noisy input latent at diffusion timestep $t$. The ACD objective is
\begin{equation}
	{L_{ACD}}=\|(\epsilon-\tfrac{\sqrt{\bar{\alpha}_{t}}}{\sqrt{1-\bar{\alpha}_{t}}}({x}^{i}-{x}^{t}))-\epsilon_\theta({x}^{i}_{t},{t}, e)\|_2.
	\label{eqn:glad_loss}
\end{equation}
where $\epsilon$ is the Gaussian noise that is added to the input feature $x^i_t$, and $alpha$ is the diffusion coefficient at timestep $t$.

This loss renders the same denoiser compatible with preservation, remove, and add pairs. Importantly, the novelty of ACD lies not in designing a new diffusion loss, but in a training strategy that constructs three input--target pairs to endow the two embeddings with complementary and decoupled semantic concepts.

\subsection{Adaptive Mask-Guided Generation}

\begin{figure*}
	\centering
	\includegraphics[pagebox=cropbox,width=.98\linewidth]{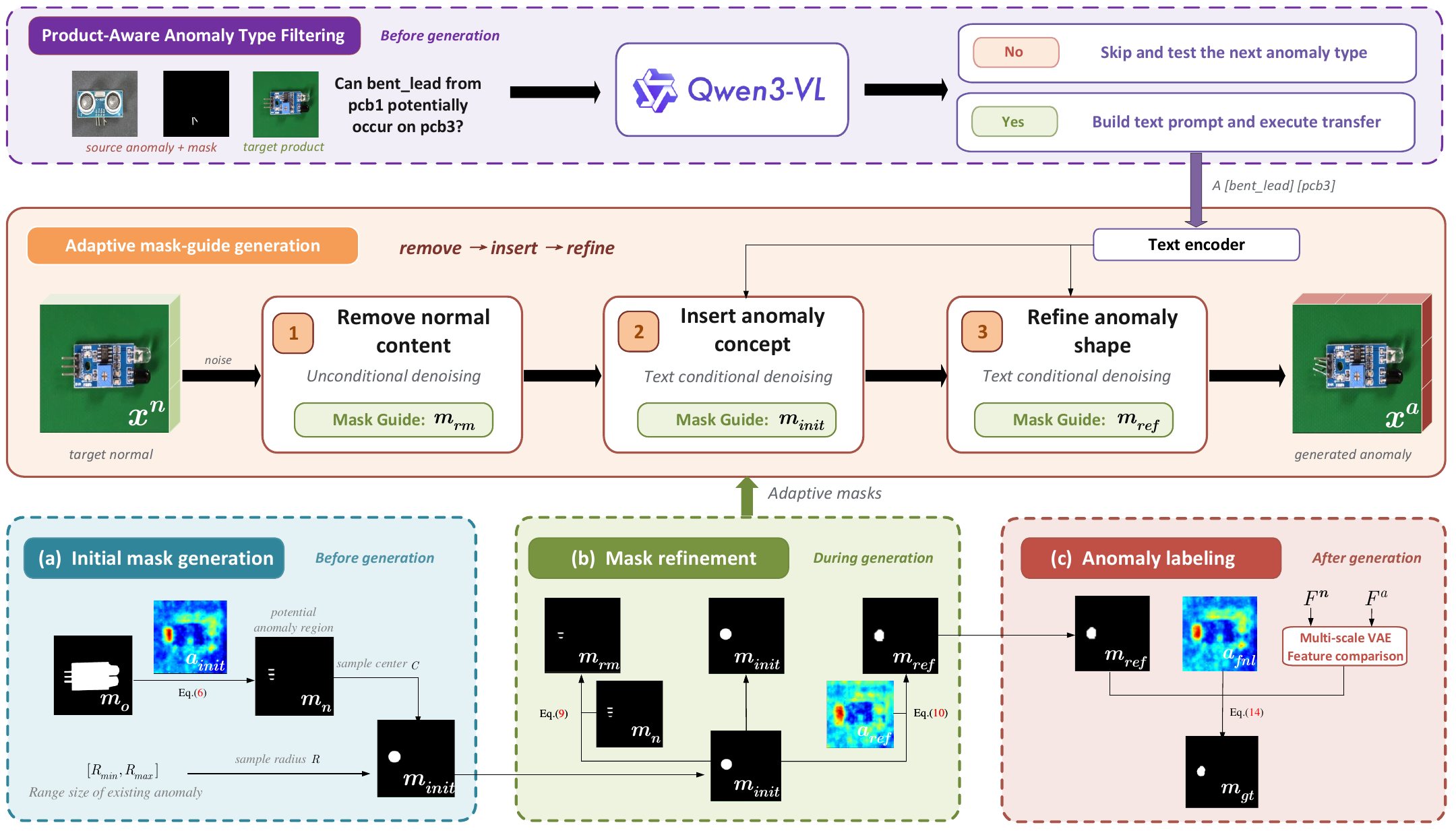}
	\vspace{-.5em}
	\caption{Overview of the cross-product anomaly transfer pipeline in DPA. The Adaptive mask-guide generation process proceeds through follows a remove--insert--refine process to generate anomaly. Each stage is controlled by a distinct text prompt and corresponding adaptive masks. For text prompt, we first filter a reasonable anomaly type and build corresponding text prompt. For adaptive masks, (a) \textbf{initial mask generation} first generates an initial mask \(m_{init}\), which provides location and size priors. (b) \textbf{mask refinement} then derives three different adaptive masks from \(m_{init}\) to govern the spatial attributes of the three stages, respectively. Finally, (b) an \textbf{anomaly labeling} module fuses refined mask, attention maps and multi-scale VAE feature differences to produce pixel-level labels \(m_{gt}\) precisely aligned with the generated anomalies.}	
	\label{fig:generation}
\end{figure*}

%
%

After obtaining the decoupled anomaly embeddings, we transfer the learned anomaly concepts to normal images of target products. As illustrated in Fig.~\ref{fig:generation}, we introduce an adaptive mask-guided process for anomaly generation. in which we control the generated content with from two aspects, and spatial attributes with adaptive masks, respectively. 

Regarding anomaly content, not every source anomaly type is applicable to given target product. Therefore, before generation, we employ a product-aware anomaly type filtering mechanism to filter anomaly type. Then, each reasonable type is combined with the target-product category to construct a text prompt, which is subsequently encoded into embeddings as the text condition.

To control the spatial attributes of the anomaly, we introduce an adaptive mask-guided process to manipulate these attributes in a stage-wise manner. There are two key modules are involved, including \textit{Initial Mask Generation} and \textit{Mask Refinement}. \textit{Initial Mask Generation} first generates initial mask for controlling plausible location and size. \textit{Mask Refinement} further refines initial mask to produce three adaptive masks for removing conflicting normal content, inserting anomaly concept and refining the anomaly shape. 

Furthermore, after generation, a \textit{Generated-Sample Verification} is utilized for discarding failed outputs.

\subsubsection{\bf Product-Aware Anomaly Type Filtering}
The availability of real source anomalies does not imply that all of them should be transferred to every target. Let $(I_s^a,m_s^a,c_s)$ denote a source anomalous image, its mask, and its anomaly type, and let $(I_p^n,c_p)$ denote a normal exemplar and the category of target product $p$. Before generation, Qwen3-VL~\cite{bai2025qwen3} receives $(I_s^a,m_s^a,c_s,I_p^n,c_p)$ and judges whether anomaly type $c_s$ could plausibly occur on product $c_p$. If the answer is negative, DPA skips this anomaly type and tests the next candidate. If the answer is positive, DPA accepts the source-anomaly--target-product pair and executes the subsequent transfer.

This compatibility decision is performed offline once for each source-anomaly-type--target-product candidate and cached for all subsequent samples. Qwen is therefore not invoked during diffusion sampling or downstream detector training and inference.

\subsubsection{\bf Initial Mask Generation}
Initial Mask Generation provides the coarse spatial constraint that determines the anomaly location and size. We construct a circular mask $m_{init}$ whose center $C$ controls location and whose radius $R$ controls size. The circle is only an initialization and its shape is refined during diffusion rather than used as the final annotation.

Anomalies usually occur in regions that are semantically relevant to the anomaly type. For example, a bent-lead anomaly can only appear in lead regions rather than arbitrary background areas. Therefore, instead of randomly sampling anomaly locations, we first identify potential anomaly regions according to the semantic relationship between the target product and the transferred anomaly concept. 

Specifically, we compute a cross-attention map $a_{init}\in\mathbb{R}^{r\times r}$ between anomaly text features $t^a\in\mathbb{R}^{Z\times C_1}$ and target-normal image features $v^n\in\mathbb{R}^{r\times r\times C_2}$ from the Stable Diffusion UNet~\cite{SD30}. The text transformer maps the anomaly embedding $e^a$ to $Z$ token features. We aggregate attention over these tokens and over UNet layers $k$ through $L$:

\begin{equation}
	a_{init}=\frac{1}{Z(L-k)}\sum_{z=0}^{Z}\sum_{l=k}^{L}\operatorname{Softmax}(\frac{Q_lK_l^\top}{\sqrt{d}}),
	\label{eq:atten}
\end{equation}

\begin{equation}
	Q=\phi_q(v^n), K=\phi_k(t^a).
\end{equation}

Here, $Q$ and $K$ are the projected query and key features, and $d$ is their channel dimension. Because ACD suppresses source-product appearance in $e^a$, high responses in $a_{init}$ indicate target regions semantically associated with the anomaly rather than regions resembling the source product.

The attention map is subsequently binarized and constrained within the object region mask $m_o$ to obtain the potential anomaly region mask $m_n$:

\begin{equation}
	m_n=\operatorname{Binarize}(m_o\cdot a_{init}).
\end{equation}

The anomaly center is randomly sampled from the potential region:

\begin{equation}
	C=Sample({C_i \mid C_i(m_n)=1}).
\end{equation}

After determining the location, we control anomaly size with statistics from the source anomaly data. We compute the range $[R_{min},R_{max}]$ of circumcircle radii and sample

\begin{equation}
	R=
	Sample([R_{min},R_{max}]).
\end{equation}

The sampled center and radius define $m_{init}\in\mathbb{R}^{r\times r}$, providing semantic location and empirical size priors while retaining diversity across generated samples.

\subsubsection{\bf Mask Refinement}
The coarse masks can only roughly constrain the position and size of generated anomalies. However, there are two finer masks are required to remove exiting normal contents and refine the shape of generated anomalies, respectively. For these two purposes, mask refinement refines the initial masks under different conditions.

First, although $m_{init}$ specifies the generation region, normal product structures may still remain within this area and interfere with subsequent anomaly generation. For example, when generating a bent-lead anomaly, the original normal lead should be removed before introducing the anomaly. Therefore, a dedicated removal mask is required to identify the normal product regions that should be erased and to guide unconditional generation for normal-content removal. We obtain such a mask $m_{rm}$ by intersecting the initial mask $m_{init}$ with the potential anomaly-prone region $m_n$:

\begin{equation}
	m_{rm}=m_{init}\cdot m_n,
\end{equation}

The removal mask $m_{rm}$ therefore erases only the normal structure that occupies the intended anomaly region, preparing it for anomaly insertion while preserving the surrounding target content.

Second, the coarse mask itself cannot accurately align the shape of the generated anomaly. Consequently, it provides only limited spatial guidance and may restrict precise shape control during the later stages of generation. To obtain a mask that better aligns with the generated anomaly, we further refine the initial mask according to the anomaly-related attention map:

\begin{equation}
	m_{ref}=\operatorname{Binarize}(m_{init}\cdot a_{ref}),
\end{equation}

where $a_{ref}$ is recomputed according to Eq.\ref{eq:atten} after anomaly content has begun to emerge. Unlike the fixed circle, $m_{ref}$ follows the current anomaly response and provides a shape-aware constraint for the final diffusion stage.

Thus, $m_{rm}$ and $m_{ref}$ serve different purposes: the former clears conflicting normal content, whereas the latter aligns subsequent generation with the emerging anomaly shape.

\subsubsection{\bf Generation with Adaptive Masks}
During generation, we start from a noisy normal latent and generate anomaly with different masks, wihch are activated at different stages of the remove--insert--refine process to progressively control anomaly generation. Specifically, the removal mask $m_{rm}$ is first used to remove residual normal content within the designated region by unconditional generation. Subsequently, under the guidance of the learned embedding, anomaly content is generated within the coarse mask $m_{init}$, which provides a reasonable constraint on anomaly location and size. Finally, the refined mask $m_{ref}$ is employed to provide more accurate shape-aware spatial constraints, enabling finer control over anomaly shapes and producing more realistic anomaly structures.

Formally, the mask-guided latent update at timestep $t$ is defined as

\begin{equation}
	x^a_t=m\cdot\hat{x}^a_t+(1-m)\cdot x^n_t,
	\label{eq:mask-replace}
\end{equation}
where $\hat{x}^a_t$ denotes the denoised latent feature at timestep $t$, and $x^n_t$ represents the noisy latent feature of the normal sample. The mask $m$ corresponds to the active adaptive mask at the current generation stage, namely $m_{rm}$, $m_{init}$, or $m_{ref}$.

Progressively switching from $m_{rm}$ to $m_{init}$ and then $m_{ref}$ mirrors the required operations: remove conflicting normal content, insert the anomaly at a plausible position and scale, and refine its final shape.

\subsubsection{\bf Generated-Sample Verification}
\label{sec:anomaly_filter}
Product-aware filtering decides whether an anomaly type is semantically suitable before generation, but diffusion may still fail to express an accepted concept in an sample. We therefore apply a separate output-level check that rejects failed samples. We define the anomaly saliency score $s$ as the difference between the mean attention inside and outside $m_{ref}$:
\begin{equation}
	s=\operatorname{Mean}(m_{ref} \cdot a_{fnl})-\operatorname{Mean}((1-m_{ref}) \cdot a_{fnl}).
\end{equation}
If $s$ is below a predefined threshold $\tau$, the generated sample is discarded. If more than one third of an inference batch fails this test, we stop the remaining generation for that source--target pair. This is an operational safeguard for synthesis failure and is distinct from the semantic compatibility decision made before transfer.

\subsection{Anomaly Labeling}

The generation masks control diffusion but are not yet suitable as segmentation labels. In particular, $m_{ref}$ is only $64\times64$ and remains a spatial constraint rather than a pixel-accurate description of the generated appearance. We therefore derive the final label from the actual change between the normal and anomalous outputs.

The VAE decoder exposes this change at progressively higher resolutions. Because mask-guided generation preserves content outside the controlled region, discrepancies between the normal and generated features are concentrated within the anomaly regions. Comparing the two multi-scale decoder features therefore complements the semantic but low-resolution attention masks.

Specifically, we consider the latent features and decoded images as the first and final layers of the decoder, denoted by $F^a = [x^a, D^a_0, D^a_1, D^a_2, \hat{I}^a]$ and $F^n = [x^n, D^n_0, D^n_1, D^n_2, \hat{I}^n]$ for the anomalous and normal samples, respectively. For each scale $d$, we compute the cosine distance between $F^a_d$ and $F^n_d$, multiply it by $m_{ref}$, normalize, and upsample to obtain multi-scale masks:
\begin{equation}
	m_{gen}^d = \operatorname{Up}(\operatorname{Norm}(m_{ref} \cdot \operatorname{Cos}(F^a_d, F^n_d))).
\end{equation}

Finally, the multi-scale masks are integrated with the attention maps $a_{fnl}$ to produce the final high-resolution ground-truth mask:
\begin{equation}
	m_{gt} = \operatorname{Binarize}(\operatorname{Norm}(\operatorname{Up}(m_{ref} \cdot a_{fnl}) + \frac{1}{D}\sum_{d=1}^{D} m_{gen}^d)).
\end{equation}

The fusion uses $m_{ref}$ to suppress irrelevant changes, $a_{fnl}$ to retain anomaly semantics, and the multi-scale feature distances to recover boundaries. The resulting $m_{gt}$ is therefore aligned with the generated anomaly rather than inherited from the initial circle.

\section{Experiments}
\label{sec:experiment}

\subsection{Datasets}We conduct experiments on two widely used industrial anomaly detection datasets, MVTec-AD~\cite{MVTec-AD28} and VisA~\cite{VisA}. To test cross-dataset transfer, each dataset serves as the source of real anomaly concepts when the other serves as the unseen target; no anomalous image from the target dataset is used for generation training.

Morerover, we further construct a dedicated benchmark, \textit{Anomaly Transfer-based Anomaly Detection (ATAD)}, to isolate the transfer setting. ATAD integrates product categories from MVTec-AD, VisA, and two in-the-wild datasets, Real-IAD~\cite{wang2024real} and MANTA~\cite{fan2025manta}. Its source and target partitions contain 13 and 16 product categories, respectively. Every target anomaly type has a semantically corresponding source type, but the source and target products differ. This correspondence is metadata inherent to the benchmark and is reported only to characterize its coverage; neither DPA nor the Qwen prompt accesses the mapping. We learn anomaly concepts only from the source partition, generate training data from normal images in the target partition, and evaluate anomaly detection on held-out target images. Supplementary material provides the category mapping and construction details.

\subsection{Setting}We evaluate two anomaly generation settings:
\subsubsection{\bf Zero-shot} source normal/anomalous images and target normal training images are available, but no target anomalous image is observed. The complete target test set is reserved for downstream evaluation.
\subsubsection{\bf Few-shot} a subset of target anomalies is additionally available. Following AnomalyDiffusion~\cite{hu2024anomalydiffusion}, we split target anomalous samples by using one third to train the generator and reserving the remaining two thirds for downstream testing.

\subsection{Implementation Details}
We use Stable Diffusion 1.5~\cite{SD30} as the generator. Only the newly introduced text-encoder embeddings are optimized; other pre-trained components remain frozen. We use Qwen3-VL-8B to filter semantically compatible source-anomaly/target-product pairs in an offline preprocessing step, and each decision is cached and reused for all images generated from that pair. Generation conditions are formed only for the accepted pairs, and Qwen is not part of per-image generation or downstream inference. For each target category, we generate 2,000 anomalous images. If $N$ anomaly types are compatible with the category, each type contributes $2000/N$ samples. We set the initial noise timestep to $T=900$ and use 10 inference steps. Unless stated otherwise, DRAEM~\cite{zavrtanik2021draem} is the common downstream detector, so differences reflect the generated training data rather than the detection architecture. All baselines use their official code and released models, and all experiments run on one NVIDIA RTX A6000 GPU. Additional hyperparameters are provided in the supplementary material.

\subsection{Evaluation Metrics}

We evaluate image-level detection and pixel-level localization with Area Under the Receiver Operating Characteristic Curve (AUROC), Average Precision (AP), and maximum F1-score (F1). The prefixes \textit{I-} and \textit{P-} denote image- and pixel-level metrics, respectively. Per-Region Overlap (PRO) additionally measures region-level localization quality. The reported mean averages the seven metrics over the evaluated datasets. More evaluation results are available in the supplementary material.


\subsection{Results of Zero-shot Anomaly Generation}
\subsubsection{\bf Comparison with Zero-shot Generation Methods}
We first compare DPA with zero-shot generation methods, including DRAEM~\cite{zavrtanik2021draem}, CPR~\cite{li2024target}, RealNet~\cite{zhang2024realnet}, GLASS~\cite{chen2024unified} and AnomalyAny~\cite{sun2025unseen}.
Here, ``DRAEM'' denotes its native generation module. Every method supplies generated image--mask pairs to the same DRAEM detector under an identical training protocol, isolating generation quality as the experimental variable.

As reported in Tab~\ref{tab:zero-shot_anomaly_generation}, DPA obtains the best overall mean of 84.91, exceeding the strongest baseline, AnomalyAny (81.44), by 3.47 points. The advantage is particularly clear for localization: relative to the best competing value, DPA improves P-AP by 8.02 points on ATAD and 8.08 points on MVTec-AD, and improves P-F1 by 3.20 points on VisA. These gains support effectiveness of DPA.

\begin{table*}[t]
	\centering
	\caption{Comparison with different zero-shot anomaly generation methods. Each method supplies generated data to the same DRAEM detector. ``DRAEM'' denotes its native generation module. Bold indicates the best result.}
	\label{tab:zero-shot_anomaly_generation}
	\resizebox{\linewidth}{!}{
		\begin{tabular}{@{}l|ccc|cccc|ccc|cccc|ccc|cccc|c@{}}
			\toprule                                  
			\multirow{2}{*}{AG Methods} & \multicolumn{7}{c|}{ATAD} & \multicolumn{7}{c|}{MVTec-AD} & \multicolumn{7}{c|}{VisA} & \multirow{2}{*}{Mean} \\
			& I-AUROC & I-AP  & I-F1  & P-AUROC & P-AP  & P-F1  & PRO   & I-AUROC & I-AP  & I-F1  & P-AUROC & P-AP  & P-F1  & PRO   & I-AUROC & I-AP  & I-F1  & P-AUROC & P-AP  & P-F1  & PRO  &  \\
			\midrule                                                                               
			DRAEM      & 93.73          & 91.32          & 86.91          & 96.54          & 47.40          & 49.23          & 85.63          & 98.72          & 99.22          & 97.61          & 97.07          & 67.01          & 65.52          & 90.73          & 96.88          & 97.49          & 94.18          & 97.33          & 29.82          & 35.96          & 86.32          & 81.17          \\
			CPR        & 94.37          & 92.41          & 87.81          & 96.16          & 44.14          & 47.25          & 83.67          & 98.82          & 99.32          & 97.76          & 96.48          & 67.27          & 65.86          & 89.54          & 96.22          & 97.00          & 92.64          & 96.30          & 29.70          & 38.12          & 83.87          & 80.70          \\
			RealNet    & 93.12          & 89.50          & 85.16          & 95.54          & 42.94          & 46.61          & 84.53          & 98.76          & 99.25          & 97.60          & 96.92          & 67.09          & 64.63          & 90.62          & 96.93          & 97.33          & 93.88          & 96.69          & 29.95          & 36.99          & 87.21          & 80.54          \\
			GLASS      &94.81           & 92.79          & 89.53          & 96.30          & 36.46          & 43.04          & 86.74      & 98.89              & 99.30 & 97.95 & 97.15 & 67.36 & 65.92 & 90.82 & 97.28 & 97.52 & 94.50 & 97.32 & 25.81 & 35.43 & 89.06 & 80.67 \\
			AnomalyAny & 93.87          & 91.30          & 86.50          & 97.81          & 42.85          & 48.19          & 88.17          & 98.60          & 98.86          & 97.86          & 97.65          & 66.95          & 64.51          & 91.25          & 96.63          & 97.50          & 92.52          & \textbf{98.50} & 32.06          & 40.15          & 88.48          & 81.44          \\
			Ours      & \textbf{96.65} & \textbf{94.66} & \textbf{90.80} & \textbf{98.89} & \textbf{55.42} & \textbf{57.00} & \textbf{91.67} & \textbf{99.25} & \textbf{99.60} & \textbf{98.34} & \textbf{98.06} & \textbf{75.44} & \textbf{71.40} & \textbf{94.10} & \textbf{98.23} & \textbf{98.50} & \textbf{95.79} & 98.47 & \textbf{38.33} & \textbf{43.35} & \textbf{89.17} & \textbf{84.91} \\
			\bottomrule
		\end{tabular}
	}
\end{table*}

\subsubsection{\bf Combination with Different Detection Frameworks}
The preceding comparison fixes the detector to isolate generation quality. We next replace the native synthesized data of DRAEM~\cite{zavrtanik2021draem}, CPR~\cite{li2024target}, and GLASS~\cite{chen2024unified} with DPA-generated data while retaining each detection framework. As shown in Tab~\ref{tab:zero-shot_anomaly_generation_baeline}, the overall mean increases from 81.17 to 84.91 for DRAEM, from 86.93 to 87.66 for CPR, and from 84.24 to 84.89 for GLASS. The consistent improvements observed across three different detection pipelines suggest that the performance gains of DPA originate from the generated anomaly data, rather than from reliance on any specific detection method.

\begin{table*}[t]
	\centering
	\caption{Combination with different synthesis-based anomaly detection methods. Bold represents optimal results.}
	\label{tab:zero-shot_anomaly_generation_baeline}
	\resizebox{\linewidth}{!}{
		\begin{tabular}{@{}l|ccc|cccc|ccc|cccc|ccc|cccc|c@{}}
			\toprule                                  
			\multirow{2}{*}{AD Methods} & \multicolumn{7}{c|}{ATAD} & \multicolumn{7}{c|}{MVTec-AD}   & \multicolumn{7}{c|}{VisA} &\multirow{2}{*}{Mean} \\
			& I-AUROC        & I-AP           & I-F1           & P-AUROC        & P-AP           & P-F1           & PRO            & I-AUROC        & I-AP           & I-F1           & P-AUROC        & P-AP           & P-F1           & PRO             & I-AUROC        & I-AP           & I-F1           & P-AUROC        & P-AP           & P-F1           & PRO        & \\
			\midrule    
			DRAEM & 93.73          & 91.32          & 86.91          & 96.54          & 47.40          & 49.23          & 85.63          & 98.72          & 99.22          & 97.61          & 97.07          & 67.01          & 65.52          & 90.73          & 96.88          & 97.49          & 94.18          & 97.33          & 29.82          & 35.96          & 86.32          & 81.17          \\
			\textit{+Ours}  & \textbf{96.65} & \textbf{94.66} & \textbf{90.80} & \textbf{98.89} & \textbf{55.42} & \textbf{57.00} & \textbf{91.67} & \textbf{99.25} & \textbf{99.60} & \textbf{98.34} & \textbf{98.06} & \textbf{75.44} & \textbf{71.40} & \textbf{94.10} & \textbf{98.23} & \textbf{98.50} & \textbf{95.79} & \textbf{98.47} & \textbf{38.33} & \textbf{43.35} & \textbf{89.17} & \textbf{84.91} \\
			\midrule   
			CPR   & 96.12          & 94.94          & 91.30          & 99.01          & 54.96          & 55.62          & 93.69          & 99.65          & 99.87          & \textbf{99.34} & 99.19          & 81.91          & 75.41          & 97.73          & 97.02          & 97.46          & 93.98          & 99.12          & 51.92          & 53.03          & 94.23          & 86.93          \\
			\textit{+Ours}  & \textbf{96.65} & \textbf{95.69} & \textbf{91.98} & \textbf{99.23} & \textbf{57.66} & \textbf{57.00} & \textbf{95.13} & \textbf{99.68} & \textbf{99.89} & \textbf{99.34} & \textbf{99.21} & \textbf{82.34} & \textbf{75.84} & \textbf{97.83} & \textbf{97.37} & \textbf{97.71} & \textbf{94.83} & \textbf{99.25} & \textbf{54.40} & \textbf{55.17} & \textbf{94.69} & \textbf{87.66} \\
			\midrule   
			GLASS & 95.98          & 93.78          & 90.43          & 99.18          & 39.17          & 49.28          & 94.61          & 99.71          & 99.89          & \textbf{99.54} & \textbf{99.11} & 69.32          & 70.96          & 95.28          & 97.78          & \textbf{98.08} & 95.14          & 98.82          & 43.72          & 47.58          & 91.77          & 84.24          \\
			\textit{+Ours}  & \textbf{96.42} & \textbf{94.85} & \textbf{90.93} & \textbf{99.32} & \textbf{41.52} & \textbf{50.89} & \textbf{95.88} & \textbf{99.73} & \textbf{99.91} & 99.33          & 99.10          & \textbf{70.47} & \textbf{71.44} & \textbf{95.41} & \textbf{97.79} & \textbf{98.08} & \textbf{95.24} & \textbf{98.85} & \textbf{45.93} & \textbf{48.21} & \textbf{93.32} & \textbf{84.89}                     \\
			\bottomrule
		\end{tabular}
	}
\end{table*}

\subsubsection{\bf Qualitative Evaluation}
\begin{figure}
	\centering
	\includegraphics[width=1.0\linewidth]{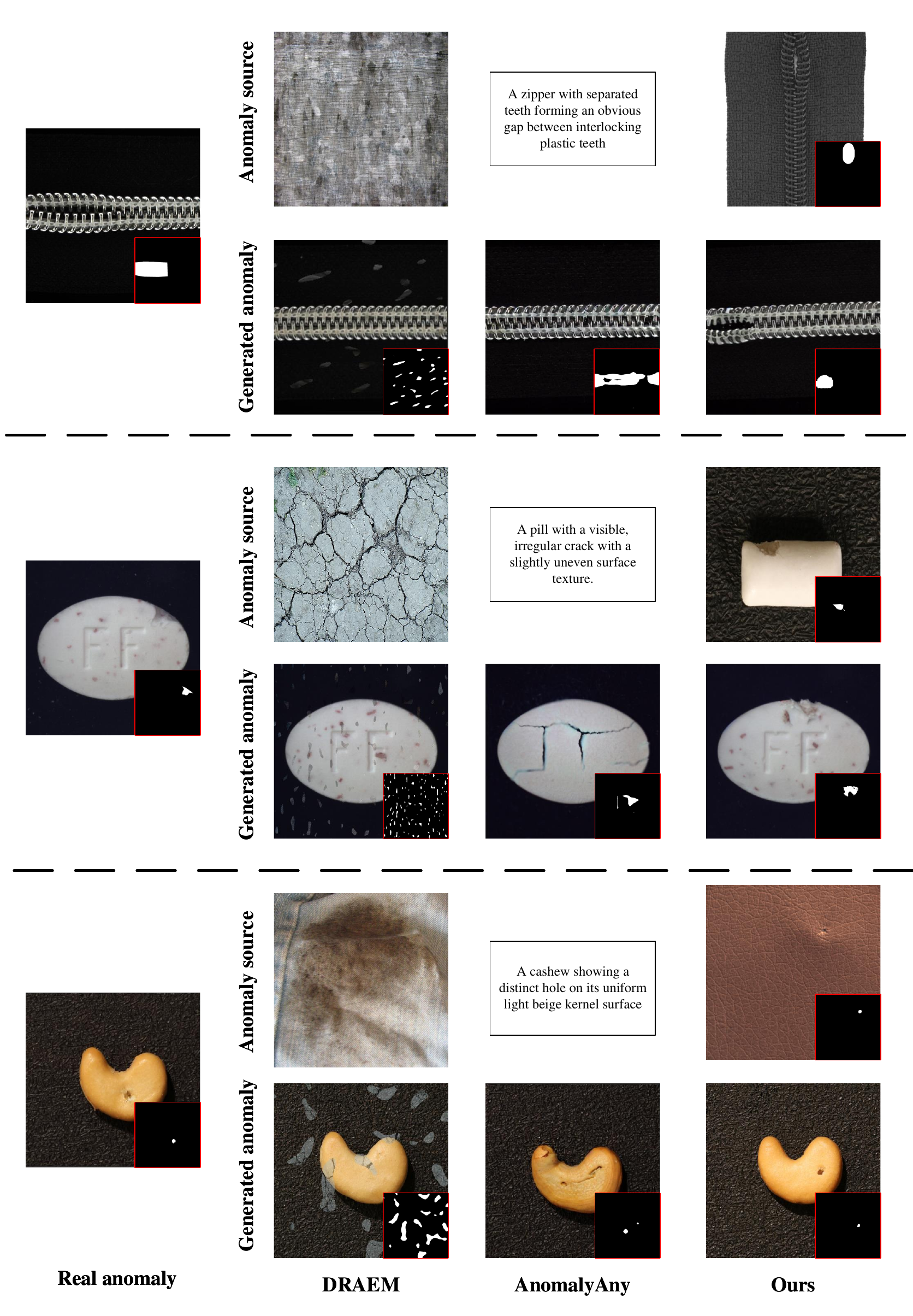}
	\caption{Qualitative comparison of different zero-shot anomaly generation methods. The first column shows real target anomalies. DRAEM uses the texture images as its anomaly source, AnomalyAny uses the displayed text descriptions, and DPA uses real source-product anomalies. Each generated image includes its predicted mask in the bottom-right inset.}
	\label{fig:zero-shot_sys}
\end{figure}

Fig.\ref{fig:zero-shot_sys} presents qualitative comparisons of different zero-shot anomaly generation methods, which use different anomaly-source. Across zipper, pill, and cashew targets, DRAEM pastes fragments of generic texture images that do not reproduce the realistic anomalies. 
AnomalyAny struggles to describe diverse abnormal appearances such as tooth separation in detail, and its understanding of 'crack' is inconsistent with the real concept of cracks. Consequently, there remains a certain gap between the generated anomalies and real-world anomalies. DPA instead conditions generation on existing source anomaly data to produce more realistic anomalies. Additional examples are provided in the supplementary material.

\subsection{Results of few-shot Anomaly Generation}

\subsubsection{\bf Performance Against Anomaly Generation Methods}
Although our method is primarily designed for zero-shot anomaly generation, it can be naturally applied to few-shot settings by using a small number of target anomaly samples for training. We compare our method with several representative few-shot generation approaches, including AnomalyDiffusion~\cite{hu2024anomalydiffusion}, AnoGen~\cite{gui2024few}, and TF-IDG~\cite{xu2025training}.

As shown in Tab~\ref{tab:few-shot}, DPA ranks first on ATAD (82.22), MVTec-AD (91.28), and VisA (82.37), yielding an overall mean of 85.29. This is 1.16 points above AnomalyDiffusion, the strongest baseline at 84.13. The same spatial-control and labeling mechanisms therefore remain useful even when the anomaly embeddings are  learned directly from target anomaly data. Thus, DPA is not limited to the zero-shot generation setting.

\begin{table}[t]
	\centering
	\caption{Comparison with different few-shot anomaly generation methods. Results for each dataset represent the mean of all metrics. Bold represents optimal results.}
	\label{tab:few-shot}
	\resizebox{0.8\linewidth}{!}{
		\begin{tabular}{@{}l|ccc|c@{}}
			\toprule                                  
			AG Methods & ATAD &MVTec-AD & VisA & Mean \\
			\midrule
			AnomalyDiffusion & 81.74          & 90.94          & 79.72          & 84.13          \\
			AnoGen           & 79.79          & 89.93          & 79.60          & 83.10          \\
			TF-IDG           & 79.10          & 88.73          & 78.41          & 82.08          \\
			Ours             & \textbf{82.22} & \textbf{91.28} & \textbf{82.37} & \textbf{85.29} \\
			\bottomrule
		\end{tabular}
	}
\end{table}

\begin{figure}
	\centering
	\includegraphics[width=1.0\linewidth]{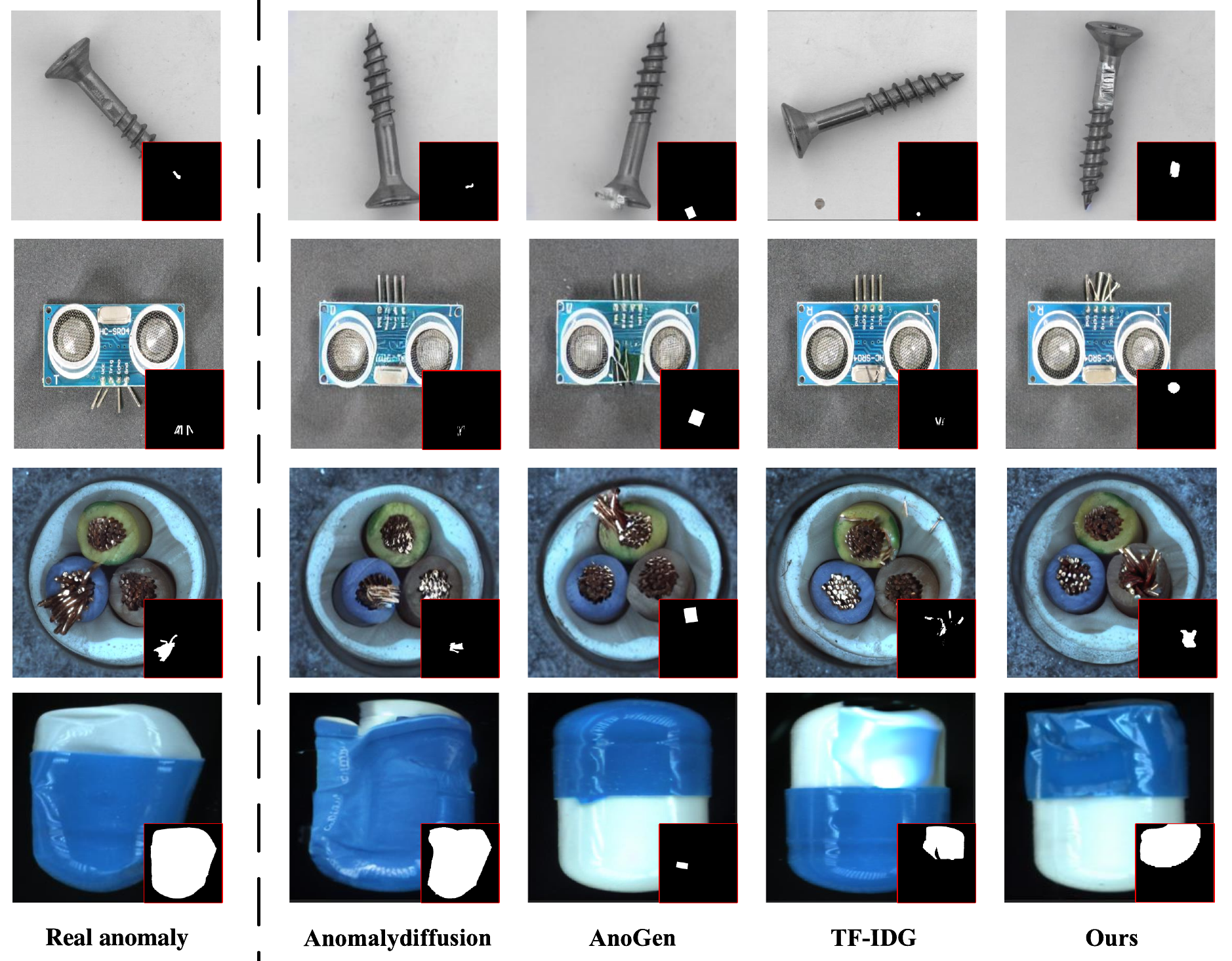}
	\caption{Qualitative comparison under the few-shot setting. The first column shows real target anomalies, and the remaining columns show results from AnomalyDiffusion, AnoGen, TF-IDG, and DPA. The bottom-right inset of each generated image is its mask.}
	\label{fig:few-shot}
\end{figure}

\subsubsection{\bf Qualitative Evaluation}
Fig.\ref{fig:few-shot} shows the generation results under the few‑shot generation setting. AnomalyDiffusion and TF-IDG can place masks on background regions, causing failures for \texttt{scratch\_neck} on \texttt{screw} and \texttt{bent\_lead} on \texttt{pcb1}. AnoGen restricts generation to the foreground product, but this constraint is insufficient, Defects such as \texttt{bent\_lead} may appear in positions without any leads. Our method produces physically plausible anomalies accompanied by well‑aligned annotations. Additional results are provided in the supplementary material.

\subsection{Ablation Study}
We now isolate the contribution of each module and test the assumptions behind cross-product transfer. Additional ablations are provided in the supplementary material.

\subsubsection{\bf Effectiveness of Each Proposed Module}
We train DRAEM on MVTec-AD with each variant's generated samples. Without ACD, anomaly embeddings are learned directly from source anomalies; without IMG, the initial circle is placed randomly; without MR, the initial mask remains fixed throughout generation; and without AL, the low-resolution $m_{ref}$ is used as the segmentation label. In Tab~\ref{tab:diffmodules}, the complete model reaches a mean of 90.88, 2.90 points above the native DRAEM-generation baseline. Removing any module reduces the performance of anoomaly detection. Each module is therefore necessary for anomaly transfer-based anomaly generation.

\begin{table}
	\centering
	\caption{Ablation of DPA modules on MVTec-AD. ``Baseline'' uses DRAEM's native generator; ``w/o'' removes the indicated DPA module. Bold denotes the best result.}
	\label{tab:diffmodules}
	\resizebox{\columnwidth}{!}{
		\begin{tabular}{@{}l|ccc|cccc|c@{}}
			\toprule
			Methods & I-AUROC       & I-AP    & I-$F_1$-max    & P-AUROC       & P-AP    & P-$F_1$-max    & PRO  &Mean\\
			\midrule
			baseline & 98.72          & 99.22          & 97.61          & 97.07          & 67.01          & 65.52          & 90.73          & 87.98          \\
			w/o ACD  & 99.18          & 99.48          & 97.91          & \textbf{98.14} & 72.97          & 69.85          & 93.67          & 90.17          \\
			w/o IMG  & 99.20          & 99.55          & 98.09          & 97.70          & 72.57          & 69.25          & 93.70          & 90.01          \\
			w/o MR   & 98.98          & 99.49          & 97.73          & 97.71          & 72.88          & 68.77          & 92.65          & 89.74          \\
			w/o AL   & 99.14          & 99.50          & 98.33          & 98.12          & 73.39          & 69.34          & 93.67          & 90.21          \\
			ours     & \textbf{99.25} & \textbf{99.60} & \textbf{98.34} & 98.06          & \textbf{75.44} & \textbf{71.40} & \textbf{94.10} & \textbf{90.88} \\
			\bottomrule
		\end{tabular}
	}
\end{table}

\subsubsection{\bf Qualitative Analysis of Anomaly Concept Decoupling}
\begin{figure}
	\centering
	\includegraphics[width=1.0\linewidth]{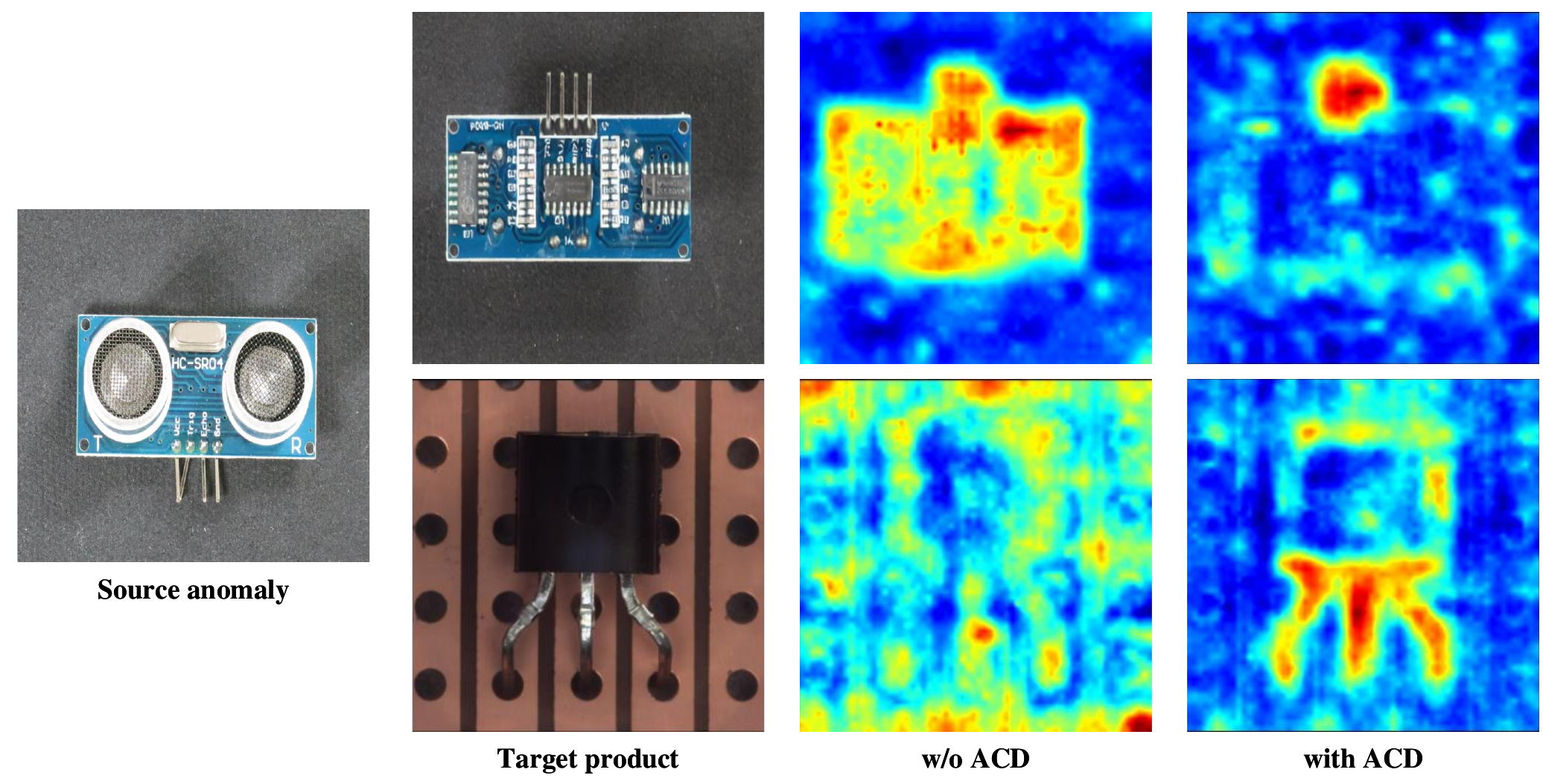}
	\caption{Cross-attention maps generated in IMG with and without ACD. After decoupling, the anomaly representation localizes semantically corresponding regions across different products, indicating a more transferable concept that is less dominated by source-product appearance.}
	\label{fig:att_location}
\end{figure}

To show the affect of ACD, Fig.\ref{fig:att_location} visualizes the cross attenion map in IMG, when transfer $\mathtt{bent\_lead}$ from source product $\mathtt{pcb1}$ to $\mathtt{pcb2}$ and $\mathtt{transistor}$. Without ACD, attention is scattered across structures resembling the source product and fails to reliably localize the target leads. With ACD enabled, the same anomaly embedding attends to the lead regions on both targets, even though the transistor differs substantially in appearance from the source PCB. This visualization thus demonstrates that our ACD helps learn anomaly concepts that are disentangled from product‑specific features.

\subsubsection{\bf Impact of Transferring Unreasonable Anomalies}
In real-world anomaly transfer scenarios, not all source anomalies are semantically compatible with a given target product. To investigate the effect of transferring such unreasonable anomalies, we consider three settings: (1) transferring only unreasonable anomaly types, which are semantically incompatible with the target product, (2) transferring only reasonable anomaly types, which are compatible and used in our default setting, and (3) transferring all anomaly types without filtering.

\begin{figure}
	\centering
	\includegraphics[width=\linewidth]{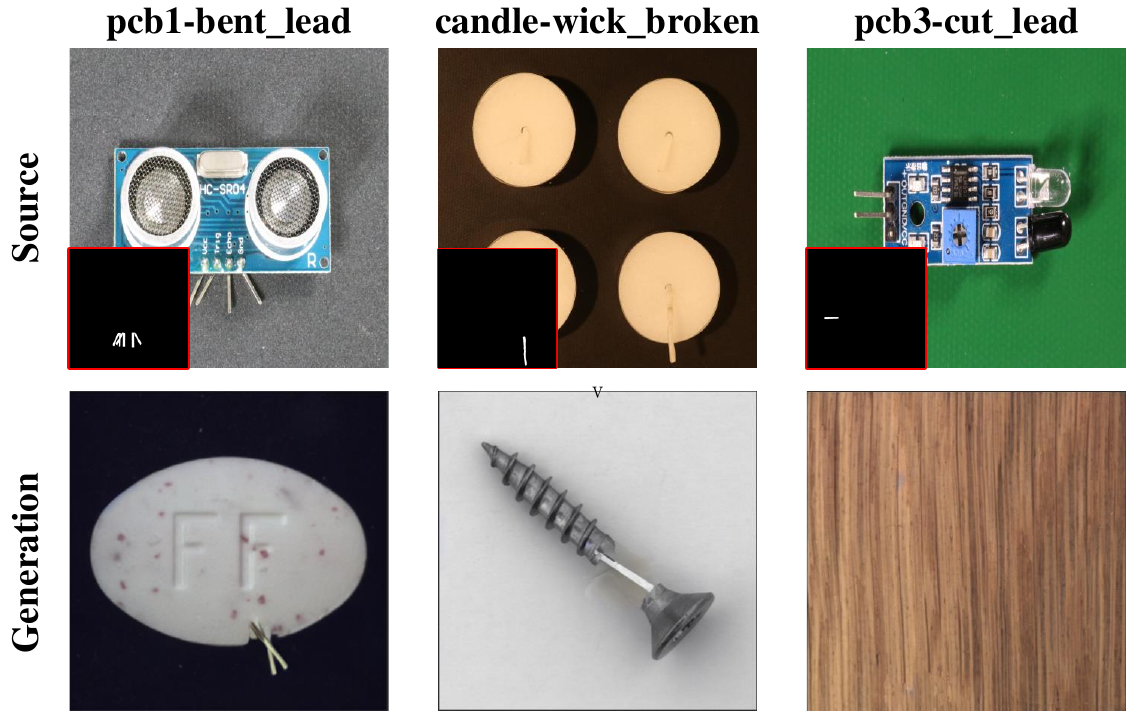}
	\caption{Results of transferring unreasonable anomalies.}
	\label{fig:unresonable}
\end{figure}

As shown in Tab~\ref{tab:diff_cover}, transferring only reasonable anomaly types yields better performance compared with transferring unreasonable anomalies. Furthermore, the intermediate result for all types also demonstrates that DPA is not catastrophically sensitive to occasional selection errors, even though reasonable types provide the optimal generated results. In addition, Fig.\ref{fig:unresonable} presents several transfer results for unreasonable anomaly types, whose generated outputs are physically unrealistic.

\begin{table}
	\centering
	\caption{Impact of transferring reasonable and unreasonable anomaly types. Bold represents optimal results.}
	\label{tab:diff_cover}
	\resizebox{\columnwidth}{!}{
		\begin{tabular}{@{}l|ccc|cccc|c@{}}
			\toprule
			transferred anomaly     & I-AUROC       & I-AP    & I-F1    & P-AUROC       & P-AP    & P-F1   & PRO  &Mean  \\
			\midrule
			Unreasonable           & 99.07          & 99.33          & 98.12          & 97.91          & 72.83          & 69.39          & 93.33          & 89.99          \\
			Reasonable type (Ours) & \textbf{99.25} & \textbf{99.60} & \textbf{98.34} & 98.06 & \textbf{75.44} & \textbf{71.40} & \textbf{94.10} & \textbf{90.88} \\
			All type               & 99.23          & 99.59          & 98.21          & \textbf{98.15}          & 73.96          & 70.25          & 93.70          & 90.44                   \\
			\bottomrule
		\end{tabular}
	}
\end{table}

\subsubsection{\bf Impact of Target Anomaly Coverage}
The proposed anomaly transfer paradigm aims to transfer generic and common anomaly types across products that are prevalent in real-world manufacturing scenarios (e.g., scratches, cracks, bent lead), rather than product-specific anomalies. Consequently, some target anomalies may not have corresponding counterparts in the source domain. To investigate the impact of such cases, we compare two experimental results: (1) transferring anomaly types that include the target anomaly category, and (2) transferring only anomaly types that exclude the target anomaly category.

As shown in Tab~\ref{tab:target_cover}, source coverage of the target anomaly type raises the performance of anomaly detection. The gap confirms that a matching source concept is useful, but the model remains competitive without one because other transferred anomalies still teach the detector to separate normal from abnormal appearance. Thus, anomaly transfer is most effective when source coverage is broad, yet it does not require a one-to-one source counterpart for every future target anomaly. If a specific uncovered type is essential, the few-shot protocol provides a natural extension.

\begin{table}
	\centering	
	\caption{Impact of target anomaly type coverage. Bold represents the best result.}
	\label{tab:target_cover}
	\resizebox{\columnwidth}{!}{
		\begin{tabular}{@{}l|ccc|cccc|c@{}}
			\toprule
			transferred anomaly     & I-AUROC       & I-AP    & I-F1    & P-AUROC       & P-AP    & P-F1   & PRO  &Mean  \\
			\midrule
			w/o target type         & 99.20          & 99.50          & 98.30          & 97.99          & 72.83          & 69.86          & 94.01          & 90.24          \\
			with target type (Ours) & \textbf{99.25} & \textbf{99.60} & \textbf{98.34} & \textbf{98.06} & \textbf{75.44} & \textbf{71.40} & \textbf{94.10} & \textbf{90.88}                   \\
			\bottomrule
		\end{tabular}
	}
\end{table}

\subsection{Computational Cost}
\begin{table}
	\caption{Train and test time cost on MVTec-AD dataset.}
	\label{tab:train_test_time}
	\resizebox{\linewidth}{!}{
		\begin{tabular}{@{}l|c|ccc@{}}
			\toprule                
			Method          & Resolution & Train/hours & Test/hours & Total/hours \\
			\midrule 
			DRAEM           & -          & -           & 0.25       & 0.25  \\
			\midrule 
			AnomalyDiffusion & 256        & 153.7       & 51.7       & 205.4 \\
			AnoGen          & 256        & 30.5        & 12.7       & 43.2  \\
			TF-IDG          & 512        & -           & 124        & 124   \\
			AnomalyAny      & 512        & -           & 291.5      & 291.5 \\
			Ours            & 512        & 70.9        & 5.5        & 76.4  \\
			\bottomrule
		\end{tabular}
	}
\end{table}

Tab~\ref{tab:train_test_time} reports the cost of generating 2,000 images for each product of MVTec-AD dataset. DPA has a total runtime of 76.4 hours, which is lower than AnomalyDiffusion (205.4 hours), TF‑IDG (124 hours) and AnomalyAny (291.5 hours). Although AnoGen is faster, it only generates images at a resolution of \(256\times256\), whereas DPA operates at \(512\times512\). Furthermore, DPA outperforms AnoGen in inference speed, which yields greater advantages when numerous generated samples are required.

\section{Conclusion}
\label{sec:conclusion}
We introduced anomaly transfer-based zero-shot generation, which converts anomaly scarcity on a new product to a cross-product transfer problem. The real anomalies from existing products can provide the visual prior, while no anomalous target image is required. Because not every anomaly is suitable for every product, DPA first uses product-aware filtering to retain only reasonable source anomaly types. It then decouples anomaly concept from source-product appearance, and generates the anomaly with adaptive masks, finally deriving a high-resolution label from the generated anomaly. Results on MVTec-AD, VisA, and ATAD show that the resulting data outperform existing zero-shot generate methods, improve multiple downstream detectors, and remain effective under the few-shot generation setting.

\bibliographystyle{IEEEtran}
\bibliography{main}

\end{document}


\title{Supplementary Material}
\markboth{Journal of \LaTeX\ Class Files,~Vol.~14, No.~8, August~2021}%
{Shell \MakeLowercase{\textit{et al.}}: A Sample Article Using IEEEtran.cls for IEEE Journals}


\maketitle

\section{Appendix}
\label{sec:appendix}

The content of this supplementary material is organized as follows:
 
\begin{itemize}
\item More implementation details in \cref{sec:implementation}.
\item Additional Metrics in \cref{sec:metrics}.
\item ATAD dataset in \cref{sec:ataddataset}.
\item Additional Ablation Study in \cref{sec:additional_ablation}.
\item Discussion on Inpainting Image Quality in \cref{sec:quality_inpainting_images}.
\item Results of Anomaly Type Filtering in \cref{sec:anomaly_filtering}.
\item Additional Results of Anomaly Generation in \cref{sec:result_generation}
\item Additional Results of Anomaly Detection in \cref{sec:result_detection}.
\item Additional Visualization in \cref{sec:visualization}.
\end{itemize}

\section{More implementation details}
\label{sec:implementation}

\subsection{More Training Details}
In our method, anomaly types and product categories are represented as bracketed text, such as ``$\mathtt{a\ [bent\_lead]}$", ``$\mathtt{a\ [pcb1]}$", and ``$\mathtt{a\ [pcb1]\ with\ [bent\_lead]}$", respectively. To learn these concepts, we adopt Textual Inversion~\cite{TextInversion35}, which adds token ids and corresponding learnable embeddings for each new word in the tokenizer and text encoder. Each product category is represented by 8 learnable embeddings, while each anomaly type is represented by 4 learnable embeddings. These embeddings are jointly optimized during training and serve as the textual representations of products and anomalies throughout the generation process.

During the decoupling training phase, the batch size is set to 6, and the learning rate is 0.032 without scaling. We train the learning embeddings for 3000 steps per anomaly type. In the zero-shot setting, we additionally train the product embeddings with the same training parameters as in the decoupling phase, except for a batch size of 16. In the few-shot setting, we directly use the trained decoupled product embeddings without further training.

For downstream anomaly detection, we directly adopt parameters of the original algorithm.

\subsection{More inference details}
During generation, we load the decoupled anomaly embeddings and the additionally trained target-product embeddings into the text encoder. The prompt combines the target product with the selected anomaly type. Starting from a noisy normal sample at timestep $T=900$, the diffusion model performs 10 denoising steps to generate an anomalous image. The adaptive mask is switched progressively during these 10 steps. Specifically, step $0$ uses the removal mask $m_{rm}$ to remove conflicting normal content, steps $1$--$6$ use the initial mask $m_{init}$ to insert the anomaly with plausible location and size, and steps $7$--$9$ use the refined mask $m_{ref}$ to refine the anomaly shape. In the Generated-Sample Verification, the anomaly saliency threshold is set to $\tau=0.05$.

\section{Additional Metrics}
\label{sec:metrics}
Besides the standard anomaly detection metrics, we additionally report Accuracy (ACC) and Intersection over Union (IoU) for methods with binary segmentation outputs of DRAEM~\cite{zavrtanik2021draem} in \cref{sec:result_detection}. These metrics are particularly relevant in practical industrial applications.

In addition, to evaluate the quality of generated anomalies, we adopt three commonly used metrics. For zero-shot generation, since there is no strict target category sample, we evaluate the Inception Score (IS). For few-shot generation, we use the commonly used LPIPS (IC-LPIPS)~\cite{ojha2021few} and Kernel Inception Distance (KID)~\cite{binkowski2018demystifying}.

\section{ATAD dataset}
\label{sec:ataddataset}
To facilitate research on anomaly transfer, we construct a new benchmark named the Anomaly Transfer-based Anomaly Detection dataset (ATAD) by integrating product categories and anomaly types from four widely used industrial anomaly detection datasets, including the manually curated datasets MVTec-AD~\cite{MVTec-AD28} and VisA~\cite{VisA}, as well as the in-the-wild datasets Real-IAD~\cite{wang2024real} and MANTA~\cite{fan2025manta}.

\begin{figure}
	\centering
	\includegraphics[width=\linewidth]{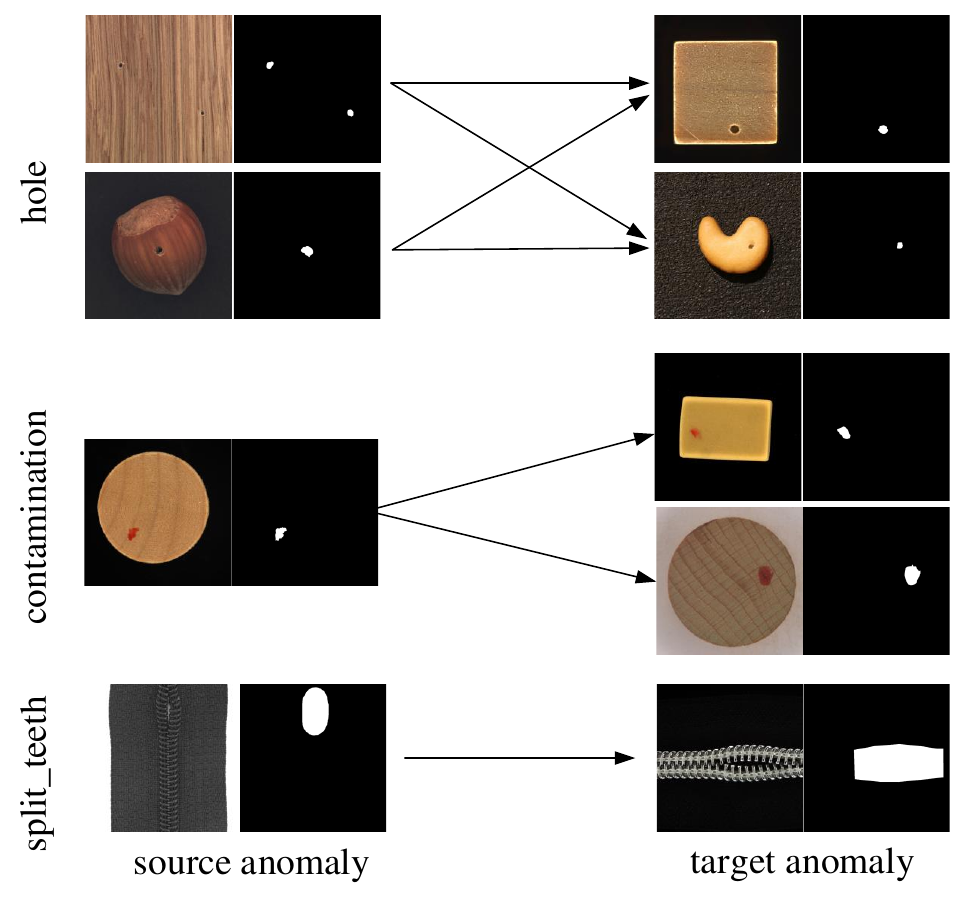}
	\caption{Similarity between source and target anomalies in ATAD dataset.}
	\label{fig:dataset}
\end{figure}

ATAD is divided into a source subset that provides reference anomalies and a target subset used for generation and detection. Each target anomaly category has at least one semantically corresponding source type. Example correspondences are shown in \cref{fig:dataset}, and the complete many-to-many mapping is provided in {\tt ATAD\_anomaly\_type\_match.csv}. The mapping is metadata inherent to the benchmark and is provided only to document its anomaly-type coverage. It is not an input to DPA, the Qwen prompt, prompt construction, training, or hyperparameter selection. Our experiments instead use Qwen3-VL-8B to select plausible source--target pairs without consulting the mapping or any target anomaly label.

Detailed dataset statistics are summarized in \cref{tab:dataset}. The source subset contains 13 product categories, including 6,308 normal training images, 2,170 normal test images, and 1,886 anomalous test images. The target subset contains 16 product categories, including 8,555 normal training images, 2,858 normal test images, and 1,993 anomalous test images. The selected products cover diverse industrial scenarios, ranging from single-instance objects to multi-instance products (e.g., $\mathtt{candles}$, $\mathtt{capsules}$, and $\mathtt{macaroni2}$) and structurally complex products (e.g., $\mathtt{pcb2}$ and $\mathtt{pcb3}$), providing a challenging and diverse benchmark for anomaly transfer research.

\begin{table}
	\centering
	\caption{Statistical overview of the ATAD dataset. 	For each category, the number of training samples, normal test samples, anomalous test samples and anomaly types is provided.}
	\label{tab:dataset}
	\resizebox{\columnwidth}{!}{
		\begin{tabular}{@{}l|c|cccc@{}}
			\toprule
			dataset                          & category         & train (normal) & test (normal) & test (anomalous) & anomaly type \\
			\midrule
			\multirow{14}{*}{source sub-dataset} & bottle\_cap     & 370   & 369        & 263           & 3                  \\
			& capsule\_mvtec  & 219   & 23         & 109           & 5                  \\
			& fire\_hood      & 418   & 418        & 169           & 4                  \\
			& hazelnut        & 391   & 40         & 70            & 4                  \\
			& macaroni1       & 900   & 100        & 100           & 5                  \\
			& oblong\_tab t  & 509   & 194        & 36            & 4                  \\
			& pcb1            & 904   & 100        & 100           & 5                  \\
			& pcb4            & 904   & 101        & 100           & 6                  \\
			& red\_washer     & 578   & 148        & 38            & 3                  \\
			& transistor1     & 266   & 265        & 469           & 3                  \\
			& usb\_adaptor    & 362   & 361        & 280           & 4                  \\
			& wood            & 247   & 19         & 49            & 4                  \\
			& zipper\_mvtec   & 240   & 32         & 103           & 6                  \\
			\cline{2-6}
			& total           & 6308  & 2170       & 1886          & 56                 \\
			\midrule
			\multirow{17}{*}{target sub-dataset} & cable           & 224   & 58         & 35            & 3                  \\
			& candle          & 900   & 100        & 90            & 4                  \\
			& capsule\_manta & 500   & 200        & 40            & 2                  \\
			& capsules        & 542   & 60         & 50            & 3                  \\
			& cashew          & 450   & 50         & 78            & 4                  \\
			& eraser          & 389   & 389        & 235           & 4                  \\
			& fryum           & 450   & 50         & 80            & 4                  \\
			& macaroni2       & 900   & 100        & 100           & 5                  \\
			& pcb2            & 901   & 100        & 86            & 4                  \\
			& pcb3            & 905   & 101        & 77            & 4                  \\
			& pill            & 267   & 26         & 96            & 4                  \\
			& plastic\_nut    & 442   & 442        & 118           & 4                  \\
			& toy\_brick      & 371   & 370        & 261           & 4                  \\
			& type\_c         & 620   & 120        & 31            & 2                  \\
			& woodstick       & 443   & 442        & 116           & 4                  \\
			& zipper\_realiad & 251   & 250        & 500           & 4                  \\
			\cline{2-6}
			& total           & 8555  & 2858       & 1993          & 59					\\
			\bottomrule
		\end{tabular}
	}
\end{table}

\section{Additional Ablation Study}
\label{sec:additional_ablation}

\noindent{\bf Influences of Threshold in Mask Binarization.}
We further investigate the impact of different threshold values in the mask binarization process during generation. The results presented in the \cref{tab:threshold} indicate that a threshold of 0.55 yields the optimal detection performance overall.

\begin{table}
	\centering
	\caption{Comparison of different thresholds. Bold represents optimal results.}
	\label{tab:threshold}
	\resizebox{\columnwidth}{!}{
		\begin{tabular}{@{}l|ccc|cccc|c@{}}
			\toprule
			Threshold & I-AUROC       & I-AP    & I-$F_1$-max    & P-AUROC       & P-AP    & P-$F_1$-max    & PRO  &Mean\\
			\midrule                          
			0.45 & 99.07          & 99.43          & 98.11          & 98.16          & 73.00          & 68.83          & 93.26          & 89.98          \\
			0.50 & 99.12          & 99.40          & 97.83          & \textbf{98.18} & 73.53          & 69.67          & 93.59          & 90.19          \\
			0.55 (Ours) & \textbf{99.25} & \textbf{99.60} & \textbf{98.34} & 98.06          & \textbf{75.44} & \textbf{71.40} & \textbf{94.10} & \textbf{90.88} \\
			0.60 & 99.13          & 99.48          & 98.10          & 97.67          & 71.18          & 68.57          & 93.22          & 89.62           \\
			\bottomrule
		\end{tabular}
	}
\end{table}

\noindent{\bf Impact of Normal Region Removal and Generated-Sample Verification.}

As shown in \cref{tab:normal_region_remove}, introducing the proposed Normal Region Removal (NRR) module and Generated-Sample Verification (GSF) module consistently improves all evaluation metrics. The NRR module removes normal content within the mask region via unconditional generation before anomaly generation, preventing interference from original structures. GSF removes generated samples that lack visible anomalies despite having anomaly labels. Without such filtering, these mislabeled normal samples would pollute the training set and harm downstream detection. The results confirm the importance of both modules for anomaly detection performance.

\begin{table}
	\centering
	\caption{Ablation of Normal Region Removal (NRR) and Generated-Sample Verification (GSF). Bold denotes the best result.}
	\label{tab:normal_region_remove}
	\resizebox{\linewidth}{!}{
		\begin{tabular}{@{}l|ccc|cccc|c@{}}
			\toprule
			& I-AUROC       & I-AP    & I-$F_1$-max    & P-AUROC       & P-AP    & P-$F_1$-max    & PRO  &Mean \\
			\midrule
			w/o GSF & 98.99          & 99.41          & 97.78          & 97.79          & 72.40          & 68.73          & 93.57          & 89.81          \\
			w/o NRR  & 99.09          & 99.46          & 98.13          & 97.90          & 73.50          & 69.36          & 93.84          & 90.18          \\
			ours    & \textbf{99.25} & \textbf{99.60} & \textbf{98.34} & \textbf{98.06} & \textbf{75.44} & \textbf{71.40} & \textbf{94.10} & \textbf{90.88} \\
			\bottomrule
		\end{tabular}
	}
\end{table}

\section{Discussion on Inpainting Image Quality}
\label{sec:quality_inpainting_images}
To rule out the possibility that poor inpainting-normal images degrade anomaly generation, we adopt two complementary measures. First, during training, we paste the original anomaly back onto the inpainted image with mask to further eliminate residual differences in the normal regions. This ensures that even if the inpainted image contains minor artifacts, they will not be mistakenly learned as anomaly patterns. Second, we evaluate the distributional fidelity of the inpainted results. Using KID and FID on MVTec‑AD, we compare the distance from inpainted images to real normal samples against that from raw anomalies to normal samples. The significantly smaller distance of inpainted images confirms that the inpainting model produces satisfying counterparts.  \cref{fig:dataset} further show no severe artifacts in the completed regions. This ensures that subsequent training benefits from reliable paired data.

\begin{table}[t]
	\centering
	\caption{Quality of the completed inpainting-normal counterparts on MVTec-AD. Lower KID and FID indicate a distribution closer to the real normal data.}
	\label{tab:inpainting_quality}
	\resizebox{\columnwidth}{!}{
		\begin{tabular}{@{}lcc@{}}
			\toprule
			Computed with real normal images & KID$\downarrow$ & FID$\downarrow$ \\
			\midrule
			real anomalous images               & 0.0956 & 114.64 \\
			inpainting-normal images            & \textbf{0.0517} & \textbf{71.83} \\
			\bottomrule
		\end{tabular}
	}
\end{table}

\begin{figure}
	\centering
	\includegraphics[width=\linewidth]{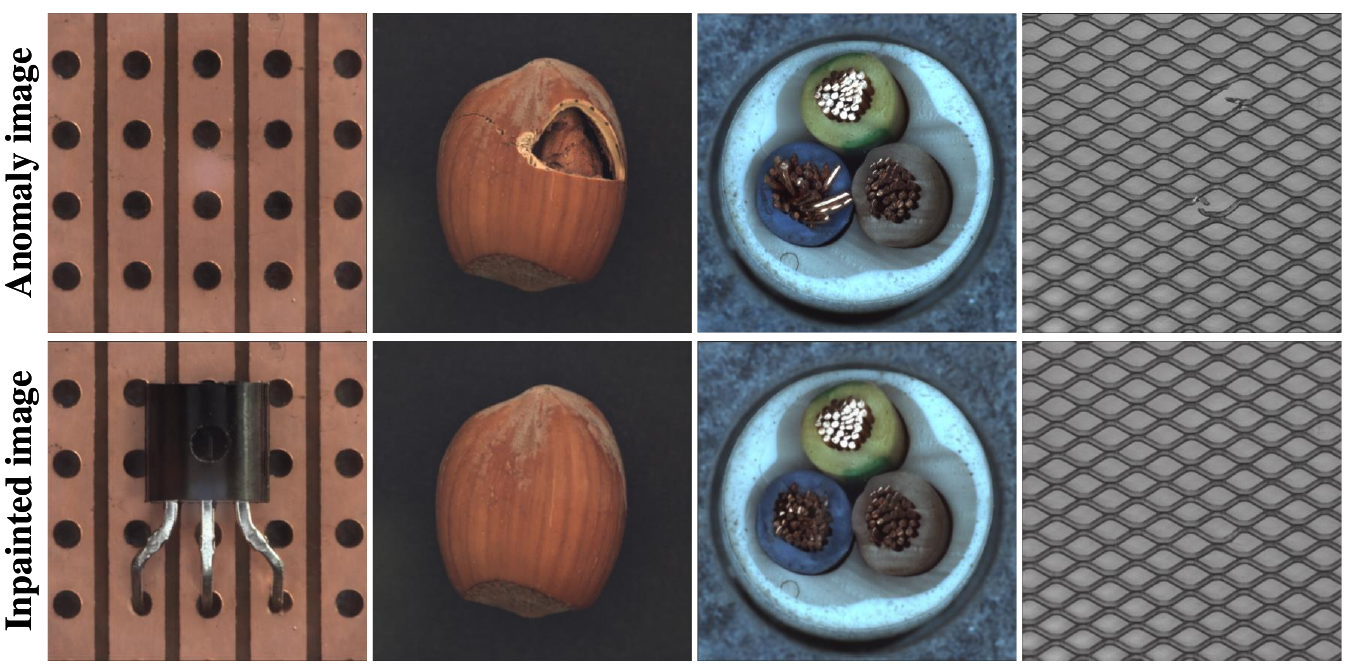}
	\caption{Qualitative examples of inpainted pseudo-normal images.}
	\label{fig:inpainting}
\end{figure}

\section{Results of Anomaly Type Filtering}
\label{sec:anomaly_filtering}
Our method transfers source anomalies to target products, which inherently involves the selection of anomaly types. To facilitate anomaly transfer, we systematically split anomaly data and renamed anomaly types. 
Comprehensive details are included in the additional supplementary materials {\tt MVTec-AD\_anomaly.csv} (for MVTec-AD dataset), {\tt VisA\_anomaly.csv} (for VisA dataset), {\tt ATAD\_source\_anomaly.csv} and {\tt ATAD\_target\_anomaly.csv} (for ATAD dataset).

Because semantically incompatible transfers can degrade detection, Qwen3-VL filters source--target pairs before generation. It receives the source anomalous image, its mask and type, together with a normal exemplar and the category of the target product, then judges compatibility under a fixed instruction. This is a one-time offline decision for each source-anomaly-type--target-product candidate. The result is cached and reused for all generated samples, so Qwen is not called during diffusion sampling or downstream detector training and inference. 
The complete filtered outputs are provided in the additional supplementary materials:{\tt VisA2MVTec-AD\_filter.csv} for MVTec-AD, {\tt MVTec-AD2VisA\_filter.csv} for VisA, and {\tt source2target\_filter.csv} for ATAD.

\section{Additional Results of Anomaly Generation}
\label{sec:result_generation}
\Cref{tab:zero-shot_is_mvtec,tab:few-shot_iclpips_mvtec} quantify generation quality. In the zero-shot generation setting, target anomaly types are unavailable during training, so generated and real defects cannot be paired by type; we therefore report IS but not IC-LPIPS or KID. In the few-shot generation setting, real target anomalies provide valid same-type references, allowing IC-LPIPS and KID to measure diversity. DPA obtains the best mean in both tab s.

\begin{table}
	\centering
	\caption{IS comparison under the zero-shot setting on MVTec-AD. Bold denotes the best result.}
	\label{tab:zero-shot_is_mvtec}
	\resizebox{0.6\linewidth}{!}{
		\begin{tabular}{@{}l|cc@{}}
			\toprule
			Class     & AnomalyAny  & DPA \\
			\midrule
			bottle     & 2.17 & 1.47 \\
			cable      & 2.95 & 1.89 \\
			capsule    & 1.71 & 1.90 \\
			carpet     & 1.24 & 1.26 \\
			grid       & 2.32 & 2.49 \\
			hazelnut   & 2.19 & 2.06 \\
			leather    & 1.37 & 3.04 \\
			metal\_nut & 1.53 & 1.53 \\
			pill       & 2.40 & 1.91 \\
			screw      & 1.20 & 1.69 \\
			tile       & 3.44 & 2.54 \\
			toothbrush & 1.59 & 1.51 \\
			transistor & 1.50 & 1.67 \\
			wood       & 1.97 & 3.02 \\
			zipper     & 1.88 & 1.75 \\
			\midrule
			mean       & 1.96 & \textbf{1.98}\\		
			\bottomrule
		\end{tabular}
		
	}
\end{table}

\begin{table}
	\centering
	\caption{IC-LPIPS and KID comparisonunder the few-shot setting on MVTec-AD dataset. \textbf{Bold} represent optimal results.}
	\label{tab:few-shot_iclpips_mvtec}
	\resizebox{\linewidth}{!}{
		\begin{tabular}{@{}l|cc|cc|cc@{}}
			\toprule
			\multirow{2}{*}{class} & \multicolumn{2}{c}{Anomalydiffusion} & \multicolumn{2}{c}{AnoGen} & \multicolumn{2}{c}{DPA} \\
			& IC-L$\uparrow$           & KID$\downarrow$            & IC-L$\uparrow$           & KID$\downarrow$           & IC-L$\uparrow$         & KID$\downarrow$          \\
			\midrule
			bottle                 & 0.15         & 0.07         & 0.12         & 0.12        & 0.15       & 0.09       \\
			cable                  & 0.38         & 0.07         & 0.36         & 0.11        & 0.38       & 0.04       \\
			capsule                & 0.16         & 0.04         & 0.17         & 0.05        & 0.19       & 0.11       \\
			carpet                 & 0.21         & 0.13         & 0.22         & 0.12        & 0.25       & 0.10       \\
			grid                   & 0.39         & 0.08         & 0.41         & 0.09        & 0.34       & 0.05       \\
			hazelnut               & 0.27         & 0.02         & 0.29         & 0.02        & 0.29       & 0.02       \\
			leather                & 0.33         & 0.15         & 0.35         & 0.13        & 0.32       & 0.10       \\
			metal\_nut             & 0.24         & 0.08         & 0.21         & 0.12        & 0.23       & 0.08       \\
			pill                   & 0.23         & 0.03         & 0.23         & 0.06        & 0.25       & 0.04       \\
			screw                  & 0.25         & 0.01         & 0.26         & 0.02        & 0.27       & 0.02       \\
			tile                   & 0.48         & 0.28         & 0.46         & 0.29        & 0.44       & 0.16       \\
			toothbrush             & 0.15         & 0.03         & 0.16         & 0.05        & 0.16       & 0.03       \\
			transistor             & 0.29         & 0.15         & 0.27         & 0.16        & 0.27       & 0.11       \\
			wood                   & 0.33         & 0.11         & 0.34         & 0.07        & 0.36       & 0.07       \\
			zipper                 & 0.23         & 0.11         & 0.23         & 0.13        & 0.20       & 0.04       \\
			\midrule
			mean                   & 0.27         & 0.09        &0.27         & 0.10        & \textbf{0.28}       & \textbf{0.07}    \\
			\bottomrule
		\end{tabular}
	}
\end{table}

\section{Additional Results of Anomaly Detection}
\label{sec:result_detection}

\noindent{\bf Deterministic Detection Performance.}
Since DRAEM produces binary segmentation outputs, its predictions can be interpreted as deterministic metrics that explicitly identify the presence and location of anomalies within input images. In practical industrial deployment, we require such deterministic outputs to make reliable decisions about whether to accept or reject tested products. As shown in \cref{tab:acc}, under both zero-shot and few-shot generation settings, our method brings significant improvements in deterministic metrics, which holds substantial practical importance for real-world industrial applications. 

\begin{table}[!t]
	\centering
	\caption{Comparison of deterministic metrics under zero-shot and few-shot settings. Bold denotes the best result.}
	\label{tab:acc}
	\resizebox{\linewidth}{!}{
		\begin{tabular}{@{}l|cc|cc|cc@{}}
			\toprule
			\multirow{2}{*}{AD Methods} & \multicolumn{2}{c|}{ATAD} &\multicolumn{2}{c|}{MVTec-AD}   & \multicolumn{2}{c}{VisA}     \\
			\cmidrule{2-7}
			& ACC         & IoU         & ACC         & IoU    & ACC         & IoU  \\			
			\midrule
			\multicolumn{7}{c}{zero-shot setting} \\
			\midrule
			DRAEM            & 64.85          & 58.62          & 68.81          & 54.14          & 56.67          & 55.14          \\
			CPR              & 66.01          & 60.08          & 60.19          & 50.28          & 55.78          & 52.06          \\
			RealNet          & 64.02          & 60.73          & 69.46          & 54.79          & 53.82          & 54.98          \\
			GLASS            & 60.09          & 55.89          & 67.08          & 55.36          & 52.17          & 52.82          \\
			AnomalyAny       & 66.47          & 57.09          & 67.14          & 49.56          & 52.47          & 49.36          \\
			Ours             & \textbf{70.15} & \textbf{61.96} & \textbf{75.45} & \textbf{59.26} & \textbf{62.93} & \textbf{57.90} \\
			\midrule
			\multicolumn{7}{c}{few-shot setting} \\
			\midrule
			AnomalyDiffusion & 73.37          & 65.69          & 74.55          & 62.87          & 62.46          & 61.44          \\
			AnoGen           & 74.07          & 66.08          & 77.09          & 61.22          & 65.68          & 61.77          \\
			TF-IDG           & 70.69          & 66.93          & 65.22          & 55.16          & 65.96          & 59.64          \\
			Ours             & \textbf{74.72} & \textbf{70.71} & \textbf{77.66} & \textbf{63.33} & \textbf{67.16} & \textbf{63.72} \\
			\bottomrule
		\end{tabular}
	}
\end{table}

\noindent{\bf Performance Against Unsupervised IAD Methods.}
The zero-shot generation setting uses no target anomalous image during training and is therefore comparable to unsupervised IAD. We additionally compare with SimpleNet~\cite{liu2023simplenet}, Dinomaly~\cite{guo2025dinomaly}, and INP-Former~\cite{luo2025exploring}. For a fair evaluation, center-crop augmentation is disabled because it can remove anomalous regions. \Cref{tab:unsupervise} shows that detection models trained with DPA-generated data remain competitive with these unsupervised approaches.

\begin{table*}[!t]
	\centering
	\caption{Comparison with SOTA unsupervised anomaly detection methods. Bold represents optimal results.}
	\label{tab:unsupervise}
	\resizebox{\linewidth}{!}{
		\begin{tabular}{@{}l|ccc|cccc|ccc|cccc|ccc|cccc|c@{}}
			\toprule
			\multirow{2}{*}{AD Methods} & \multicolumn{7}{c|}{ATAD} &\multicolumn{7}{c|}{MVTec-AD}   & \multicolumn{7}{c|}{VisA}     & \multirow{2}{*}{Mean} \\
			& I-AUROC & I-AP  & I-F1  & P-AUROC & P-AP  & P-F1  & PRO   & I-AUROC & I-AP  & I-F1  & P-AUROC & P-AP  & P-F1  & PRO  & I-AUROC & I-AP  & I-F1  & P-AUROC & P-AP  & P-F1  & PRO &  \\
			\midrule
			
			DRAEM      & 93.73          & 91.32          & 86.91          & 96.54          & 47.40          & 49.23          & 85.63          & 98.72          & 99.22          & 97.61          & 97.07          & 67.01          & 65.52          & 90.73          & 96.88          & 97.49          & 94.18          & 97.33          & 29.82          & 35.96          & 86.32          & 81.17                    \\
			SimpleNet  & 92.34          & 91.07          & 85.24          & 98.26          & 30.89          & 36.62          & 93.32          & 98.92          & 99.55          & 98.86          & 97.93          & 51.94          & 54.33          & 92.25          & 96.16          & 96.91          & 92.64          & 98.46          & 33.16          & 37.11          & 92.29          & 79.44                    \\
			CPR        & 96.12          & 94.94          & 91.30          & 99.01          & 54.96          & 55.62          & 93.69          & 99.65          & 99.87          & 99.34 & 99.19          & 81.91          & 75.41          & 97.73          & 97.23          & 97.62          & 94.71          & 99.13          & 51.46          & 52.99          & 93.89          & 86.94                    \\
			GLASS & 95.98          & 93.78          & 90.43          & 99.18          & 39.17          & 49.28          & 94.61          & 99.71          & 99.89          & \textbf{99.54} & 99.11 & 69.32          & 70.96          & 95.28          & 97.78          & 98.08 & 95.14          & 98.82          & 43.72          & 47.58          & 91.77          & 84.24          \\
			Dinamaly   & 96.63          & 95.25          & 91.60          & 99.30          & 47.63          & 51.37          & \textbf{96.29} & 99.69          & 99.87          & 99.21          & 98.55          & 66.80          & 67.51          & 95.37          & \textbf{98.54} & \textbf{98.79} & 95.48          & 99.10          & 49.24          & 52.88          & \textbf{94.78}          & 85.42                    \\
			INP-former & 96.50          & 94.81          & 91.51          & 99.11          & 53.01          & 55.54          & 95.69          & 99.47          & 99.82          & 98.98          & 98.52          & 68.23          & 67.95          & 96.38          & 98.02          & 98.29          & 95.06          & 98.60          & 49.61          & 52.73          & 93.71          & 85.79                    \\
			\midrule
			DRAEM+Ours & \textbf{96.65} & 94.66          & 90.80          & 98.89          & 55.42          & \textbf{57.00} & 91.67          & 99.25          & 99.60          & 98.34 & 98.06          & 75.44          & 71.40          & 94.10          & 98.23          & 98.50          & \textbf{95.79} & 98.47          & 38.33          & 43.35          & 89.17          & 84.91                    \\
			CPR+Ours   & \textbf{96.65} & \textbf{95.69} & \textbf{91.98} & 99.23          & \textbf{57.66} & \textbf{57.00} & 95.13          & 99.68          & 99.89          & 99.34 & \textbf{99.21} & \textbf{82.34} & \textbf{75.84} & \textbf{97.83} & 97.37          & 97.71          & 94.83          & \textbf{99.25} & \textbf{54.40} & \textbf{55.17} & 94.69 & \textbf{87.66}           \\
			GLASS+Ours & 96.42          & 94.85          & 90.93          & \textbf{99.32} & 41.52          & 50.89          & 95.88          & \textbf{99.73} & \textbf{99.91} & 99.33          & 99.10          & 70.47          & 71.44          & 95.41          & 97.79          & 98.08          & 95.24          & 98.85          & 45.93          & 48.21          & 93.32          & 84.89                             \\
			\bottomrule
		\end{tabular}
	}
\end{table*}

\noindent{\bf Detailed Detection and Segmentation Results.}
In addition, we also report the detailed detection and segmentation results for each category on the MVTec-AD, VisA and ATAD datasets under zero-shot setting. \cref{tab:zero-shot_atad}, \cref{tab:zero-shot_mvtec} and \cref{tab:zero-shot_visa} to presents the results of two representative methods (DRAEM~\cite{zavrtanik2021draem} and AnomalyAny~\cite{sun2025unseen}) and our proposed approach.

\begin{table*}[!t]
	\centering
	\caption{Comparison under the zero-shot generation setting with DRAEM on ATAD datasets. Bold represent optimal results.}
	\label{tab:zero-shot_atad}
	\resizebox{\linewidth}{!}{
		\begin{tabular}{@{}l|cccc|ccccc|cccc|ccccc|cccc|ccccc@{}}
			\toprule
			\multirow{2}{*}{Class} 
			& \multicolumn{9}{c|}{DREAM}  
			& \multicolumn{9}{c|}{AnomalyAny}   
			& \multicolumn{9}{c}{DPA}                                                     \\
			& I-AUROC & I-AP   & I-F1   & ACC   & P-AUROC & P-AP  & P-F1  & IOU   & PRO   
			& I-AUROC & I-AP   & I-F1   & ACC   & P-AUROC & P-AP  & P-F1  & IOU   & PRO   
			& I-AUROC & I-AP   & I-F1   & ACC   & P-AUROC & P-AP  & P-F1  & IOU   & PRO   \\
			\midrule
			cable           & 94.19  & 92.90  & 87.88  & 78.49  & 98.14 & 78.76 & 70.66 & 54.86 & 86.69 & 94.83  & 93.18  & 85.71  & 83.87 & 98.16 & 78.72 & 69.63 & 60.37 & 87.46 & 98.87  & 98.31  & 94.44  & 91.40 & 99.62 & 88.11 & 78.63 & 55.19 & 94.27 \\
			candle          & 88.69  & 87.52  & 81.65  & 53.68  & 93.06 & 18.38 & 28.50 & 31.35 & 80.34 & 96.37  & 96.47  & 89.73  & 53.16 & 97.05 & 19.51 & 28.40 & 55.19 & 89.89 & 97.82  & 97.35  & 93.55  & 72.63 & 99.12 & 31.78 & 37.88 & 60.96 & 84.43 \\
			capsules        & 91.60  & 93.04  & 85.98  & 56.36  & 94.89 & 24.61 & 30.28 & 55.61 & 88.49 & 92.43  & 92.51  & 85.11  & 70.91 & 99.57 & 40.00 & 53.52 & 61.26 & 93.35 & 95.90  & 95.58  & 90.20  & 70.91 & 99.73 & 73.18 & 75.98 & 65.25 & 95.54 \\
			capsule\_mantan & 100.00 & 100.00 & 100.00 & 99.58  & 91.47 & 53.61 & 56.62 & 79.23 & 82.79 & 100.00 & 100.00 & 100.00 & 99.58 & 98.17 & 67.83 & 79.20 & 52.03 & 94.87 & 100.00 & 100.00 & 100.00 & 88.75 & 98.09 & 80.86 & 75.76 & 28.36 & 91.38 \\
			cashew          & 94.36  & 96.33  & 91.03  & 39.06  & 92.13 & 32.55 & 42.42 & 43.78 & 79.72 & 95.38  & 96.98  & 92.68  & 40.62 & 96.27 & 5.25  & 12.45 & 36.67 & 82.46 & 97.56  & 98.33  & 94.41  & 39.84 & 97.66 & 40.34 & 43.08 & 44.72 & 86.83 \\
			eraser          & 86.25  & 86.76  & 80.82  & 62.98  & 98.35 & 49.51 & 54.62 & 65.46 & 85.75 & 90.09  & 89.54  & 80.86  & 63.78 & 97.19 & 51.60 & 54.23 & 71.24 & 81.53 & 91.80  & 91.19  & 83.49  & 64.58 & 99.18 & 60.07 & 58.80 & 72.65 & 91.31 \\
			fryum           & 96.00  & 97.16  & 94.61  & 40.00  & 94.29 & 27.33 & 37.97 & 42.31 & 64.99 & 94.95  & 96.05  & 91.61  & 41.54 & 98.09 & 36.31 & 44.69 & 48.79 & 81.92 & 99.48  & 99.66  & 98.16  & 60.00 & 99.23 & 53.90 & 52.37 & 44.01 & 72.76 \\
			macaroni2       & 95.80  & 96.52  & 89.36  & 50.00  & 98.30 & 9.13  & 22.87 & 67.89 & 91.80 & 92.54  & 92.64  & 85.45  & 50.00 & 99.69 & 19.08 & 28.49 & 56.48 & 95.82 & 98.44  & 98.40  & 94.69  & 50.00 & 99.92 & 12.67 & 25.31 & 59.63 & 98.02 \\
			pcb2            & 98.21  & 97.80  & 93.92  & 68.28  & 83.91 & 0.93  & 3.83  & 56.91 & 60.01 & 98.45  & 98.11  & 94.32  & 74.73 & 95.57 & 19.20 & 23.97 & 47.41 & 84.39 & 98.62  & 98.10  & 96.09  & 61.29 & 98.96 & 29.42 & 37.26 & 65.14 & 86.55 \\
			pcb3            & 97.40  & 96.07  & 94.74  & 56.74  & 94.27 & 23.93 & 38.81 & 56.97 & 88.71 & 99.13  & 98.72  & 96.20  & 58.43 & 98.54 & 22.53 & 36.18 & 66.71 & 90.30 & 98.95  & 98.61  & 93.24  & 58.99 & 99.06 & 9.78  & 21.77 & 68.34 & 89.81 \\
			pill            & 97.40  & 99.33  & 95.79  & 29.51  & 96.21 & 47.57 & 49.36 & 35.26 & 91.71 & 96.31  & 99.00  & 96.34  & 40.16 & 98.55 & 61.47 & 61.70 & 39.60 & 95.27 & 98.84  & 99.70  & 97.94  & 63.11 & 99.64 & 78.37 & 74.27 & 58.44 & 97.43 \\
			plastic\_nut    & 87.81  & 70.37  & 67.20  & 79.29  & 99.22 & 52.04 & 51.07 & 80.04 & 85.78 & 84.29  & 61.16  & 59.53  & 80.00 & 99.53 & 50.12 & 50.46 & 57.08 & 91.02 & 91.33  & 76.60  & 71.89  & 80.00 & 99.87 & 70.17 & 65.13 & 81.02 & 93.89 \\
			toy\_brick      & 75.65  & 73.61  & 67.06  & 60.54  & 98.19 & 50.47 & 51.32 & 60.36 & 82.30 & 80.34  & 80.42  & 69.94  & 58.64 & 98.11 & 43.41 & 49.16 & 57.49 & 76.29 & 87.64  & 87.78  & 78.08  & 69.57 & 97.24 & 63.42 & 62.34 & 64.73 & 81.14 \\
			type\_c         & 91.59  & 83.83  & 80.65  & 84.11  & 90.57 & 20.80 & 29.86 & 63.17 & 69.55 & 93.17  & 79.78  & 77.97  & 81.46 & 92.25 & 22.16 & 32.56 & 76.79 & 75.75 & 98.17  & 91.77  & 86.96  & 91.39 & 98.42 & 42.70 & 46.21 & 79.13 & 92.71 \\
			woodstick       & 94.05  & 83.41  & 77.45  & 82.08  & 99.76 & 22.30 & 36.09 & 80.89 & 95.87 & 93.64  & 86.30  & 78.97  & 82.97 & 99.54 & 86.97 & 81.84 & 72.88 & 93.40 & 95.04  & 88.18  & 82.46  & 84.05 & 99.83 & 89.48 & 83.92 & 86.31 & 96.15 \\
			zipper\_realiad & 99.97  & 99.99  & 99.60  & 54.13  & 93.79 & 49.04 & 49.59 & 56.34 & 93.45 & 99.98  & 99.99  & 99.60  & 83.60 & 98.65 & 61.44 & 64.63 & 53.50 & 97.00 & 99.72  & 99.87  & 98.39  & 75.87 & 97.69 & 72.40 & 70.16 & 57.50 & 96.23 \\
			\midrule
			mean            & 93.06  & 90.92  & 86.73  & 62.18  & 94.78 & 35.06 & 40.87 & 58.15 & 82.99 & 93.87  & 91.30  & 86.50  & 66.47 & 97.81 & 42.85 & 48.19 & 57.09 & 88.17 & \textbf{96.76}  & \textbf{94.96}  & \textbf{90.87}  & \textbf{70.15} & \textbf{98.95} & \textbf{56.04} & \textbf{56.80} & \textbf{61.96} & \textbf{90.53} \\
			\bottomrule
		\end{tabular}
	}
\end{table*}
	
\begin{table*}[!t]
	\centering
	\caption{Comparison under the zero-shot generation setting with DRAEM on MVTec-AD datasets. Bold represent optimal results.}
	\label{tab:zero-shot_mvtec}
	\resizebox{\linewidth}{!}{
		\begin{tabular}{@{}l|cccc|ccccc|cccc|ccccc|cccc|ccccc@{}}
			\toprule
			\multirow{2}{*}{Class} & \multicolumn{9}{c|}{DRAEM}  & \multicolumn{9}{c|}{AnomalyAny}   & \multicolumn{9}{c}{DPA} \\
			& I-AUROC & I-AP & I-F1 & ACC & P-AUROC & P-AP & P-F1 & IOU & PRO & I-AUROC & I-AP & I-F1 & ACC & P-AUROC & P-AP & P-F1 & IOU & PRO & I-AUROC & I-AP & I-F1 & ACC & P-AUROC & P-AP & P-F1 & IOU & PRO \\
			\midrule
			capsule      & 98.68 & 99.74 & 98.17 & 29.55 & 96.78 & 48.10 & 49.56 & 42.06 & 93.04 & 97.61 & 99.50 & 97.30 & 24.24 & 98.70 & 52.79 & 52.09 & 36.11 & 92.65 & 97.53 & 99.52 & 96.74 & 28.03 & 98.37 & 58.63 & 54.82 & 37.63 & 93.66 \\
			screw        & 96.41 & 98.58 & 96.33 & 27.50 & 98.44 & 48.72 & 49.88 & 33.89 & 91.37 & 100.00 & 100.00 & 100.00 & 31.25 & 97.94 & 23.26 & 31.44 & 24.31 & 88.37 & 98.77 & 99.56 & 97.52 & 56.25 & 99.41 & 63.92 & 61.77 & 55.22 & 95.19 \\
			carpet       & 98.72 & 99.62 & 97.14 & 77.78 & 97.88 & 68.76 & 65.06 & 53.87 & 92.31 & 99.32 & 99.79 & 97.70 & 77.78 & 98.50 & 71.22 & 67.42 & 55.60 & 93.96 & 99.08 & 99.73 & 97.73 & 74.36 & 99.19 & 78.59 & 72.43 & 56.09 & 95.12 \\
			cable        & 95.65 & 97.38 & 92.06 & 87.33 & 95.05 & 61.21 & 56.61 & 39.52 & 81.22 & 94.40 & 96.43 & 90.62 & 70.67 & 94.68 & 62.66 & 57.54 & 45.62 & 81.89 & 97.32 & 98.50 & 93.92 & 92.67 & 97.13 & 71.37 & 66.34 & 46.11 & 90.59 \\
			pill         & 97.95 & 99.62 & 97.53 & 34.73 & 98.25 & 57.31 & 71.39 & 31.47 & 88.96 & 97.44 & 99.53 & 97.51 & 41.32 & 99.21 & 85.84 & 80.85 & 36.07 & 94.47 & 99.51 & 99.91 & 98.58 & 52.10 & 99.13 & 81.02 & 74.47 & 47.12 & 95.49 \\
			transistor   & 93.92 & 93.45 & 84.62 & 78.00 & 80.96 & 35.26 & 39.40 & 59.06 & 72.00 & 90.79 & 87.87 & 88.61 & 79.00 & 88.14 & 39.09 & 45.11 & 62.69 & 69.18 & 97.33 & 96.86 & 92.11 & 88.00 & 85.18 & 43.89 & 46.82 & 67.67 & 79.14 \\
			bottle       & 99.76 & 99.93 & 99.20 & 95.18 & 98.36 & 85.49 & 78.04 & 63.24 & 93.72 & 99.92 & 99.98 & 99.21 & 90.36 & 97.63 & 80.36 & 72.14 & 46.94 & 91.76 & 99.92 & 99.98 & 99.21 & 93.98 & 99.11 & 87.37 & 79.30 & 60.92 & 94.65 \\
			hazelnut     & 100.00 & 100.00 & 100.00 & 66.36 & 99.27 & 77.19 & 70.80 & 56.74 & 95.64 & 99.68 & 99.82 & 97.90 & 69.09 & 99.49 & 82.38 & 74.62 & 48.80 & 96.12 & 100.00 & 100.00 & 100.00 & 80.00 & 99.84 & 92.96 & 86.58 & 74.84 & 96.58 \\
			metal\_nut    & 100.00 & 100.00 & 100.00 & 95.65 & 99.52 & 95.51 & 90.70 & 77.74 & 96.74 & 99.90 & 99.98 & 99.47 & 94.78 & 97.05 & 78.73 & 73.28 & 53.97 & 94.98 & 100.00 & 100.00 & 100.00 & 94.78 & 98.73 & 91.53 & 86.01 & 69.69 & 96.31 \\
			toothbrush   & 100.00 & 100.00 & 100.00 & 76.19 & 99.03 & 53.70 & 55.70 & 56.92 & 93.10 & 100.00 & 100.00 & 100.00 & 78.57 & 98.61 & 44.14 & 50.83 & 50.89 & 90.14 & 100.00 & 100.00 & 100.00 & 83.33 & 99.14 & 60.54 & 60.05 & 57.35 & 91.17 \\
			grid         & 100.00 & 100.00 & 100.00 & 50.00 & 99.74 & 66.00 & 64.43 & 64.59 & 97.83 & 100.00 & 100.00 & 100.00 & 61.54 & 99.71 & 66.42 & 63.37 & 62.95 & 97.27 & 100.00 & 100.00 & 100.00 & 62.82 & 99.79 & 67.03 & 66.06 & 65.47 & 98.05 \\
			leather      & 100.00 & 100.00 & 100.00 & 75.00 & 98.82 & 61.45 & 60.24 & 60.60 & 95.03 & 100.00 & 100.00 & 100.00 & 54.03 & 99.15 & 64.93 & 65.22 & 61.18 & 95.29 & 100.00 & 100.00 & 100.00 & 79.03 & 99.58 & 68.35 & 65.77 & 62.71 & 97.59 \\
			tile         & 100.00 & 100.00 & 100.00 & 89.74 & 99.74 & 97.24 & 91.17 & 78.77 & 96.83 & 100.00 & 100.00 & 100.00 & 88.89 & 99.68 & 97.20 & 91.18 & 77.40 & 97.55 & 100.00 & 100.00 & 100.00 & 89.74 & 99.67 & 96.83 & 90.96 & 78.12 & 96.85 \\
			zipper       & 100.00 & 100.00 & 100.00 & 93.38 & 98.98 & 78.32 & 73.71 & 49.33 & 92.63 & 99.92 & 99.98 & 99.58 & 86.09 & 99.11 & 78.57 & 73.58 & 36.93 & 95.05 & 100.00 & 100.00 & 100.00 & 93.38 & 99.52 & 83.11 & 77.39 & 59.67 & 96.95 \\
			wood         & 99.82 & 99.95 & 99.16 & 55.70 & 95.26 & 70.89 & 66.15 & 44.30 & 80.54 & 100.00 & 100.00 & 100.00 & 59.49 & 97.15 & 76.69 & 69.04 & 43.96 & 90.09 & 100.00 & 100.00 & 100.00 & 63.29 & 97.53 & 81.04 & 74.67 & 50.23 & 90.48 \\
			\midrule
			mean         & 98.73 & 99.22 & 97.61 & 68.81 & 97.07 & 67.01 & 65.52 & 54.14 & 90.73 & 98.60 & 98.86 & 97.86 & 67.14 & 97.65 & 66.95 & 64.51 & 49.56 & 91.25 & \textbf{99.30} & \textbf{99.60} & \textbf{98.39} & \textbf{75.45} & \textbf{98.09} & \textbf{75.08} & \textbf{70.90} & \textbf{59.26} & \textbf{93.86} \\
			\bottomrule
		\end{tabular}
	}
\end{table*}

\begin{table*}[!t]
	\centering
	\caption{Comparison under the zero-shot generation setting with DRAEM on VisA datasets. Bold represent optimal results.}
	\label{tab:zero-shot_visa}
	\resizebox{\linewidth}{!}{
		\begin{tabular}{@{}l|cccc|ccccc|cccc|ccccc|cccc|ccccc@{}}
			\toprule
			\multirow{2}{*}{Class} & \multicolumn{9}{c|}{DRAEM}  & \multicolumn{9}{c|}{AnomalyAny}   & \multicolumn{9}{c}{DPA} \\
			& I-AUROC & I-AP & I-F1 & ACC & P-AUROC & P-AP & P-F1 & IOU & PRO & I-AUROC & I-AP & I-F1 & ACC & P-AUROC & P-AP & P-F1 & IOU & PRO & I-AUROC & I-AP & I-F1 & ACC & P-AUROC & P-AP & P-F1 & IOU & PRO \\
			\midrule
			candle       & 95.23 & 96.19 & 90.43 & 53.50 & 95.95 & 32.44 & 38.60 & 61.34 & 89.06 & 96.15 & 96.49 & 89.55 & 50.50 & 96.95 & 18.77 & 27.76 & 52.42 & 89.16 & 97.01 & 97.53 & 92.86 & 63.50 & 99.38 & 35.49 & 39.71 & 56.88 & 91.19 \\
			capsules     & 96.57 & 97.75 & 93.75 & 54.37 & 98.60 & 34.15 & 41.55 & 45.62 & 87.33 & 95.15 & 97.21 & 91.92 & 62.50 & 99.56 & 44.16 & 52.35 & 52.19 & 95.11 & 98.05 & 98.78 & 96.00 & 62.50 & 99.81 & 56.95 & 65.87 & 58.29 & 94.46 \\
			cashew       & 92.78 & 96.07 & 91.08 & 69.33 & 92.15 & 22.43 & 26.37 & 12.49 & 77.37 & 95.42 & 97.65 & 93.20 & 38.00 & 98.27 & 44.89 & 48.21 & 32.49 & 82.07 & 98.90 & 99.42 & 97.56 & 94.67 & 95.08 & 24.69 & 25.15 & 29.34 & 85.70 \\
			chewinggum   & 98.62 & 99.41 & 96.48 & 68.00 & 98.61 & 35.59 & 48.17 & 58.03 & 79.56 & 95.24 & 97.89 & 91.19 & 48.67 & 98.32 & 37.94 & 45.47 & 45.84 & 75.81 & 98.92 & 99.50 & 96.97 & 69.33 & 99.25 & 39.55 & 50.20 & 67.46 & 79.14 \\
			fryum        & 92.28 & 96.52 & 89.95 & 45.33 & 94.20 & 38.08 & 40.96 & 48.76 & 77.14 & 95.96 & 97.70 & 93.33 & 36.67 & 98.93 & 62.46 & 64.53 & 42.37 & 81.96 & 98.66 & 99.33 & 96.52 & 58.00 & 94.99 & 40.09 & 45.10 & 53.57 & 91.19 \\
			macaroni1    & 98.77 & 98.71 & 96.52 & 51.50 & 99.46 & 23.10 & 26.60 & 75.32 & 92.60 & 97.80 & 97.66 & 92.38 & 51.50 & 99.80 & 11.78 & 24.73 & 59.17 & 94.41 & 99.55 & 99.53 & 97.54 & 50.50 & 99.96 & 23.08 & 32.44 & 66.52 & 96.63 \\
			macaroni2    & 99.50 & 99.50 & 96.62 & 50.00 & 99.95 & 12.50 & 20.12 & 71.08 & 97.16 & 95.77 & 96.49 & 89.36 & 50.00 & 98.29 & 9.12 & 22.85 & 67.87 & 91.80 & 99.23 & 99.29 & 95.61 & 50.00 & 99.93 & 14.89 & 24.06 & 59.26 & 97.24 \\
			pcb1         & 96.47 & 96.51 & 91.08 & 61.50 & 99.61 & 70.03 & 68.25 & 58.21 & 88.67 & 97.46 & 97.63 & 91.54 & 63.50 & 98.89 & 21.64 & 34.50 & 46.57 & 90.29 & 99.02 & 98.91 & 96.55 & 62.50 & 99.55 & 66.62 & 61.98 & 61.48 & 89.39 \\
			pcb2         & 99.33 & 99.32 & 97.03 & 60.00 & 98.37 & 7.68  & 16.43 & 57.53 & 73.79 & 98.38 & 98.31 & 94.58 & 74.00 & 98.26 & 23.36 & 30.62 & 44.58 & 82.65 & 99.40 & 99.40 & 96.62 & 61.50 & 98.46 & 14.25 & 21.56 & 60.43 & 82.13 \\
			pcb3         & 98.23 & 96.30 & 96.04 & 57.21 & 96.87 & 16.80 & 25.80 & 66.54 & 89.25 & 98.95 & 98.84 & 96.12 & 56.22 & 97.27 & 29.09 & 41.74 & 62.75 & 90.08 & 99.02 & 98.80 & 96.62 & 58.21 & 98.62 & 30.82 & 37.05 & 64.14 & 91.00 \\
			pcb4         & 98.80 & 95.61 & 97.46 & 70.65 & 98.74 & 43.25 & 47.91 & 61.60 & 93.18 & 99.55 & 99.52 & 97.09 & 62.69 & 98.83 & 34.18 & 39.38 & 57.84 & 93.54 & 98.51 & 95.34 & 96.48 & 83.08 & 98.95 & 45.18 & 52.06 & 62.78 & 92.76 \\
			pipe\_fryum   & 95.92 & 97.97 & 93.66 & 38.67 & 95.43 & 21.83 & 30.82 & 45.22 & 90.75 & 96.94 & 98.43 & 93.84 & 35.33 & 97.22 & 37.41 & 44.03 & 39.60 & 90.89 & 96.84 & 98.31 & 95.00 & 41.33 & 96.72 & 48.80 & 51.94 & 54.60 & 93.74 \\
			\midrule
			mean         & 96.88 & 97.49 & 94.18 & 56.67 & 97.33 & 29.82 & 35.96 & 55.14 & 86.32 & 96.63 & 97.50 & 92.52 & 52.47 & 98.50 & 32.06 & 40.15 & 49.36 & 88.48 & \textbf{98.59} & \textbf{98.68} & \textbf{96.19} & \textbf{62.93} & \textbf{98.39} & \textbf{36.70} & \textbf{42.26} & \textbf{57.90} & \textbf{90.38} \\
			\bottomrule
		\end{tabular}
	}
\end{table*}

\section{Additional Visualization}
\label{sec:visualization}

\begin{figure*}
	\centering
	\includegraphics[width=\linewidth]{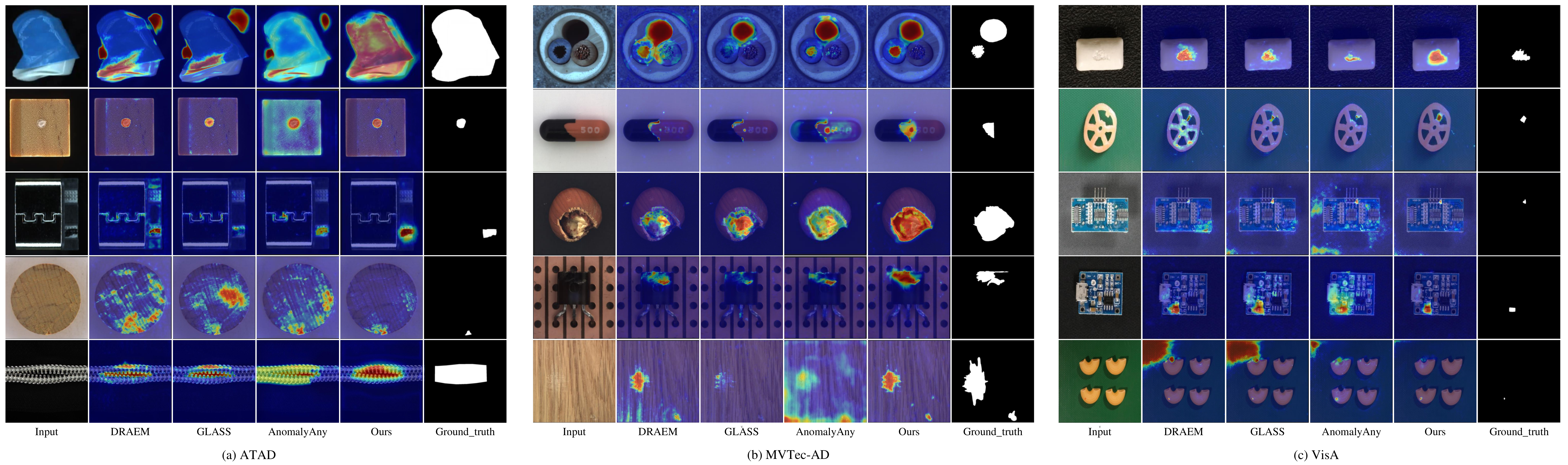}
	\caption{Anomaly localization results of DRAEM, which is trained with data from different generation methods under zero-shot generation setting. Each group shows the input anomaly, the predictions, and the ground truth.}
	\label{fig:more-zero-shot}
\end{figure*}

\begin{figure*}
	\centering
	\includegraphics[width=\linewidth]{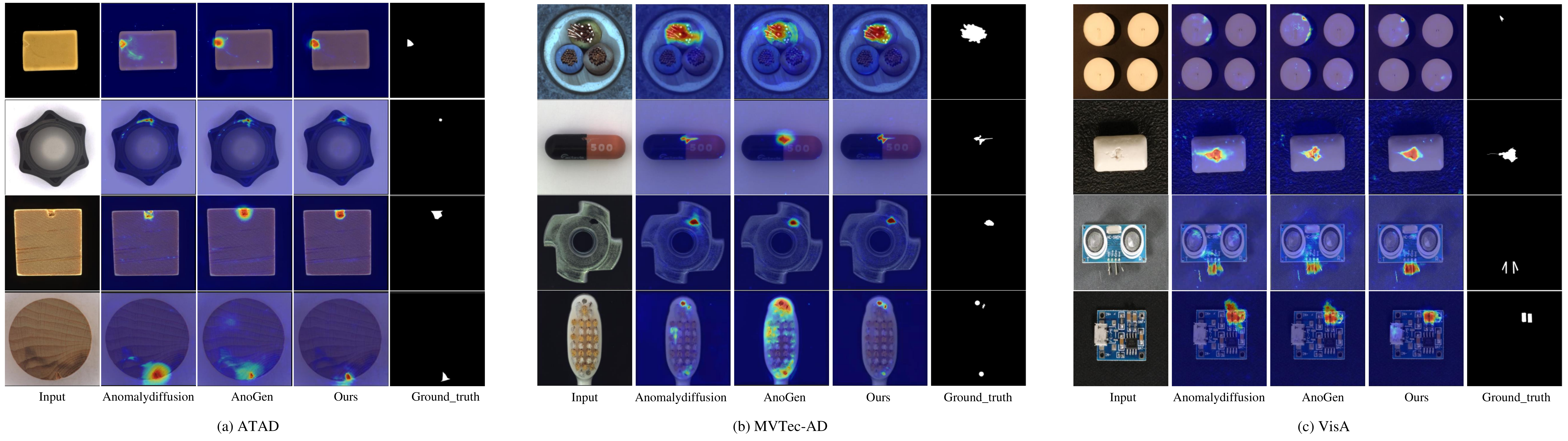}
	\caption{Anomaly localization results of DRAEM, which is trained with data from different generation methods under few-shot generation setting. Each group shows the input anomaly, the predictions, and the ground truth.}
	\label{fig:more-few-shot}
\end{figure*}

\Cref{fig:more-zero-shot,fig:more-few-shot} compare localization under the zero-shot and few-shot protocols, respectively. In every case, the same DRAEM anomaly detection model is trained with data produced by the listed generation methods, so differences reflect the generated images and labels.

\begin{figure*}
	\centering
	\includegraphics[width=0.8\linewidth]{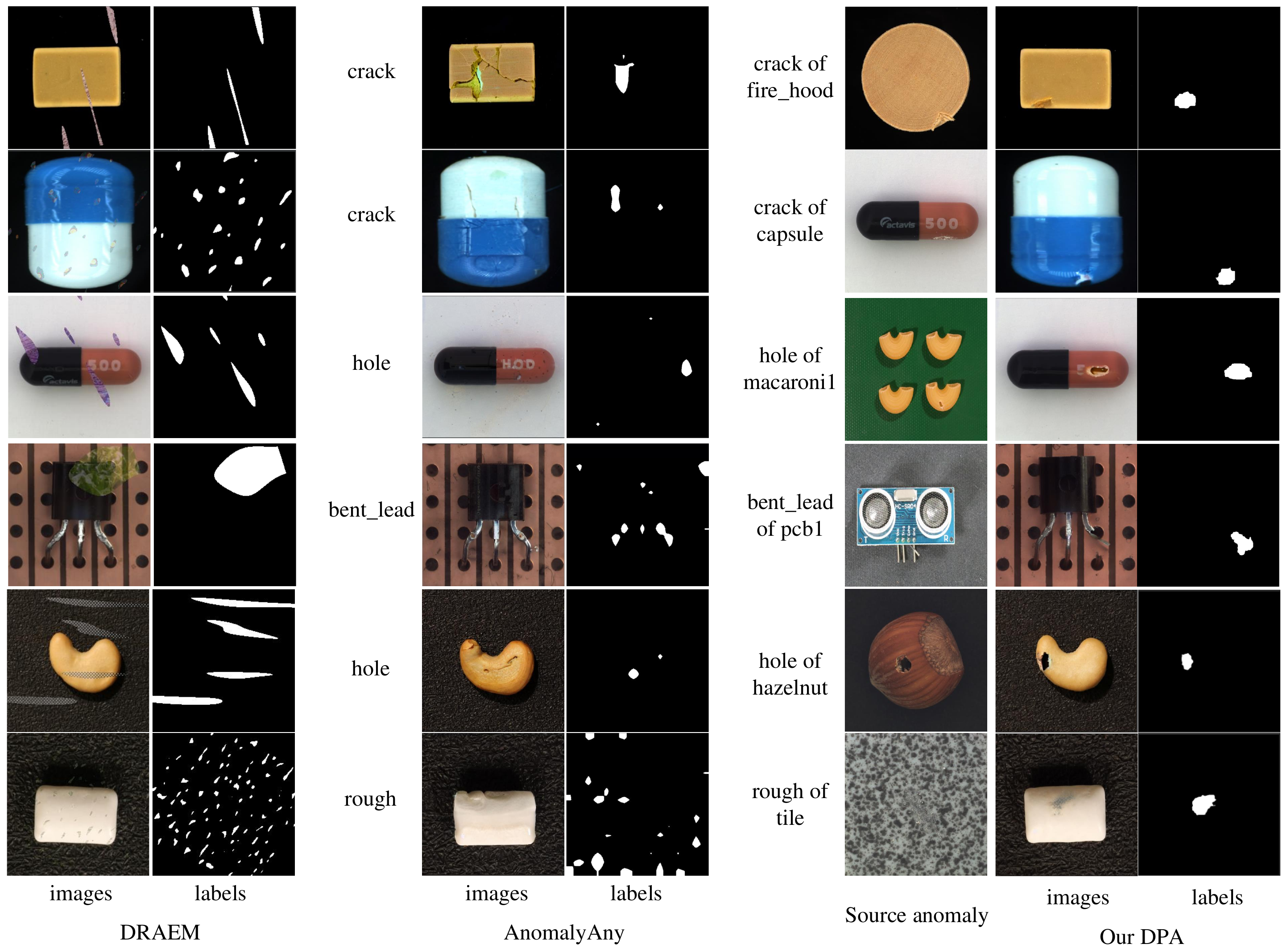}
	\caption{Additional zero-shot anomaly generation results for DRAEM, AnomalyAny, and DPA.}
	\label{fig:zero-shot_generation}
\end{figure*}

\begin{figure*}
	\centering
	\includegraphics[width=0.8\linewidth]{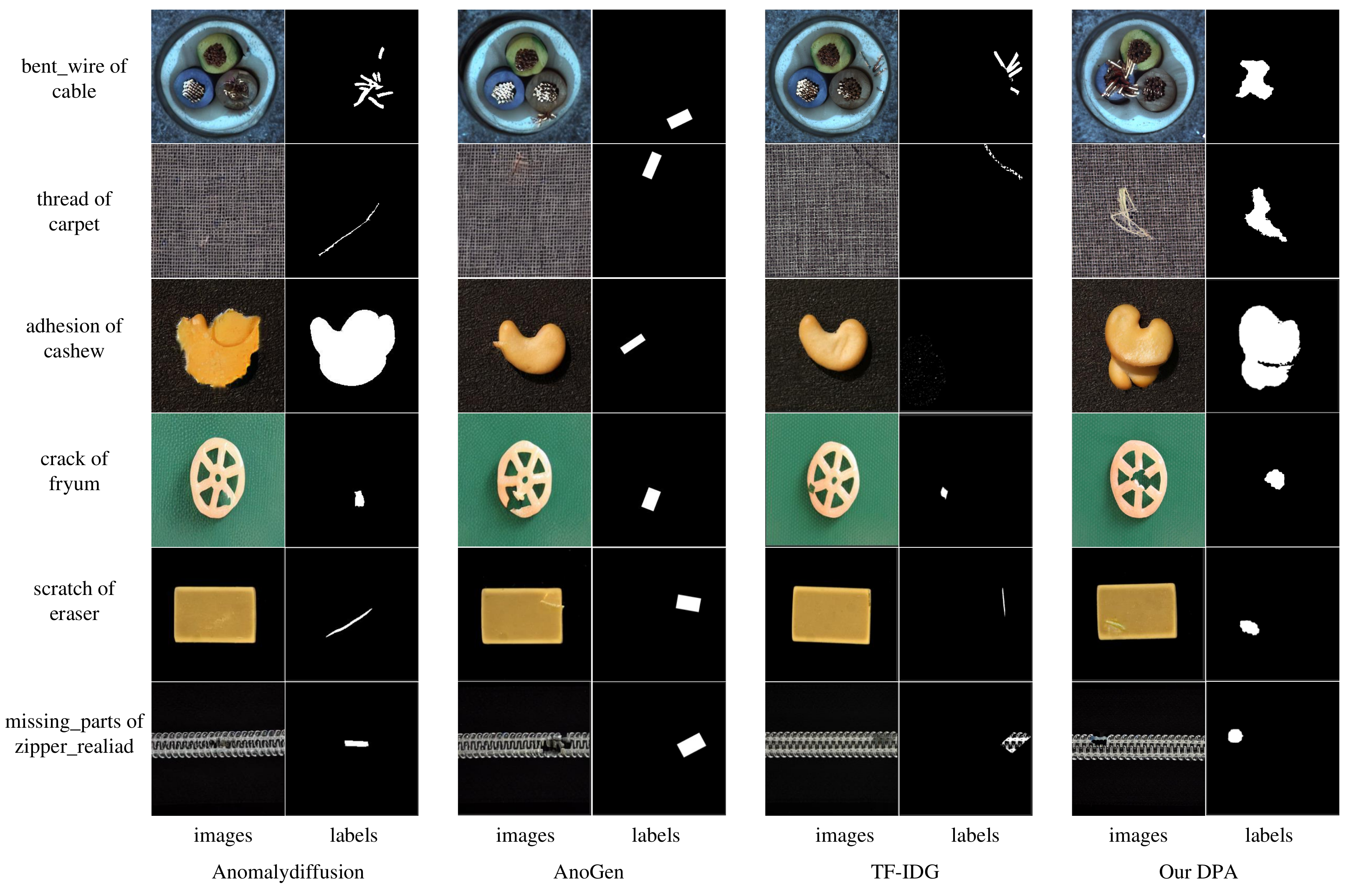}
	\caption{Additional few-shot generation results for AnomalyDiffusion, AnoGen, TF-IDG, and DPA.}
	\label{fig:few-shot_generation}
\end{figure*}

Additional zero-shot generation results are presented in \cref{fig:zero-shot_generation}, while few-shot generation results are provided in \cref{fig:few-shot_generation}.

\bibliographystyle{IEEEtran}
\bibliography{main}